\pdfoutput=1
\documentclass[journal]{IEEEtran}
\usepackage{cite}
\usepackage{amsmath,amssymb,amsfonts}
\usepackage{algorithmic}
\usepackage{graphicx}
\usepackage{textcomp}
\usepackage{booktabs}
\usepackage{colortbl}
\usepackage{multirow}
\usepackage{xcolor}
\usepackage[colorlinks=true,linkcolor=blue!50!black,citecolor=blue!50!black,urlcolor=blue!50!black]{hyperref}
\usepackage{xurl}
\usepackage{enumitem}
\graphicspath{{Images/}{Figures/}}

\hypersetup{
  pdftitle={Pothole Risk Assessment with Geometrically Grounded Vision Language Models},
  pdfauthor={Abdalmalek Aburaddaha, Zaid A. El Shair, Ali Al-Bustami, Samir A. Rawashdeh},
  pdfkeywords={autonomous vehicles, pothole detection, risk assessment, vision-language models, warning generation}
}

\begin{document}

\title{Pothole Risk Assessment with Geometrically\\ Grounded Vision Language Models}

\author{Abdalmalek~Aburaddaha,
        Zaid~A.~El~Shair,
        Ali~Al-Bustami,
        and~Samir~A.~Rawashdeh,~\IEEEmembership{Senior~Member,~IEEE}%
\thanks{This work has been submitted to the IEEE for possible publication.
Copyright may be transferred without notice, after which this version may no
longer be accessible.}%
\thanks{The authors are with the Department of Electrical and Computer
Engineering, University of Michigan--Dearborn, Dearborn, MI 48128 USA
(e-mail: abdmalek@umich.edu; zelshair@umich.edu; abustami@umich.edu;
srawa@umich.edu).}%
\thanks{Corresponding author: Abdalmalek Aburaddaha (e-mail: abdmalek@umich.edu).}}

\markboth{Preprint}%
{Aburaddaha \MakeLowercase{\textit{et al.}}: Pothole Risk Assessment with Geometrically Grounded VLMs}

\maketitle

\begin{abstract}
Potholes impose billions of dollars in annual vehicle damage, yet detectors
report only where potholes are, not how severe they are or what to do. Judging severity from a single RGB image is
ill-posed, and ungrounded vision-language model (VLM) assessments tend to be
over-conservative. We
present a five-stage modular framework for the offline assessment of
already-detected potholes that combines geometric measurements with VLM
reasoning: perspective transformation, zero-shot SAM segmentation,
geometric feature extraction, Gemini~2.5~Pro severity classification with
composite risk scoring, and natural-language warning generation. We evaluate it
on 3,092~potholes in 1,119~dashcam scenes against independent references. Severity agreement with a prediction-free three-annotator panel is
weak (Cohen's $\kappa = 0.139$, 95\%~CI $[0.00, 0.28]$; $n = 101$) and far
below the annotators' agreement with one another ($\kappa = 0.605$); against
physically measured depth on an independent RGB-D dataset it is also weak
($\kappa = 0.098$), although the predicted classes preserve the order of
measured depths. The severity classes are therefore perceived visual
categories, not depth measurements. Geometric grounding does change the
warnings: relative to an image-only baseline it moves 26.7\% of scenes out of
the high-urgency tier and raises critical-tier assignments by 109\%
(Stuart--Maxwell $p < 10^{-15}$), and in a blind
study three raters preferred the grounded warnings 3.5:1. A six-arm prompt
ablation attributes the hybrid prompt's rule-adherence to the
per-pothole measurements rather than to the rubric or worked examples. At \$0.105 per scene it suits large-scale \emph{offline}
infrastructure assessment; its maneuver suggestions
are advisory research outputs, not validated vehicle guidance.

\end{abstract}


\section{Introduction}
\label{sec:introduction}

\subsection{Motivation}
\label{sec:motivation}

\IEEEPARstart{P}{otholes} represent a critical challenge for road infrastructure and
vehicle safety, imposing substantial economic and safety burdens on transportation
systems worldwide. In the United States, drivers paid an estimated
\$26.5~billion for pothole-related vehicle repairs in 2021, at an average of
almost \$600 per repair, and about one in ten drivers needed a repair after
hitting a pothole that year~\cite{AAA2022}.

For autonomous vehicles~(AVs) and advanced driver assistance systems~(ADAS),
potholes pose unique and compounding challenges. Unlike human drivers who can
visually assess pothole severity and adjust behavior accordingly, current AV
perception systems face three fundamental limitations. First, hazard detection
remains difficult under varied conditions such as shadows, wet surfaces, and
changing illumination~\cite{Ma2022review, Strehlow2024}. Second, even when detected, existing
systems lack severity assessment capabilities; binary classification (pothole
present/absent) provides insufficient information for trajectory planning, as a
shallow surface defect requires a vastly different vehicle response than a deep
cavity risking suspension damage~\cite{Ma2022review}. Third, without actionable
guidance mechanisms, AVs cannot translate hazard detection into context-aware
maneuvers, since determining whether to brake, swerve, or proceed requires
integrating pothole depth, spatial location relative to wheel paths, and
surrounding traffic context.

Current detection methods focus predominantly on localization and segmentation
using convolutional neural networks~(CNNs) such as YOLO and Faster R-CNN,
achieving strong accuracy~\cite{ORDDC2024}. However, these approaches operate as
binary classifiers without severity quantification or actionable driver
guidance~\cite{Ma2022review, Singh2024pothole}. Vision-language models~(VLMs) offer
promising semantic reasoning capabilities, yet monocular depth estimation from
single RGB images is inherently ill-posed~\cite{Eigen2014, Saxena2005}, causing
VLMs to produce over-conservative warnings when visual evidence alone is
ambiguous. Without geometric measurements to anchor their assessments, VLMs
default to generic recommendations insufficient for severity-aware decision-making.

Instead, we propose a hybrid approach that combines VLM visual reasoning with
deterministic geometric measurements to produce severity assessments and
warnings grounded in measured image geometry. The framework assumes that
potholes have already been detected: it takes pothole bounding boxes as input
and assesses each one. It processes recorded dashcam footage offline, producing
structured risk scores and natural-language guidance intended for
infrastructure monitoring and fleet-based road assessment.

\subsection{Proposed Framework}
\label{sec:approach}

The proposed system is a five-stage modular pipeline for offline pothole hazard
assessment applied to recorded dashcam imagery~\cite{AbuRaddaha2024}. The stages
are: (1)~perspective transformation of dashcam images to reduce foreshortening and
emphasize the road surface, (2)~zero-shot pothole segmentation using
SAM~\cite{Kirillov2023}, (3)~deterministic geometric feature extraction computing
area, circularity, solidity, aspect ratio, perimeter, and lateral
distance (wheel-path intersection) from segmentation masks,
(4)~VLM-based severity classification using
Gemini~2.5~Pro~\cite{GoogleAI2025Gemini25} grounded in the extracted geometric
features, with multi-factor risk scoring combining severity, area, and spatial
position, and (5)~natural language warning generation producing
driver-interpretable maneuver recommendations. The three severity classes are
named after the depth bands low~($<$2\,cm), medium~(2--5\,cm), and
high~($>$5\,cm), following the depth-based convention of pavement maintenance
manuals~\cite{FHWA2013DistressManual, TxDOT2020, OntarioReg2002}; because no
depth is measured, they are perceived visual-severity classes rather than
measurements. The input bounding boxes are dataset annotations that stand in
for an upstream detector, and detection itself is not evaluated. Each stage
feeds structured outputs to the next, enabling systematic validation and
component-level ablation. Section~\ref{sec:methodology} provides full
implementation details.

\subsection{Contributions}
\label{sec:contributions}

This work makes the following contributions:

\begin{enumerate}

\item \textbf{A five-stage hybrid CV+VLM framework} for offline pothole risk
assessment that unifies perspective transformation, zero-shot segmentation,
deterministic geometric feature extraction, VLM-based severity classification
with composite risk scoring, and natural language warning generation into a single
modular pipeline. To our knowledge, no prior system combines deterministic
geometric measurements with VLM reasoning for pothole severity assessment; the
contribution lies in this integration and its evaluation rather than in a new
learning algorithm.

\item \textbf{An extended annotation resource (ENPAD)} built upon an existing
public dataset~\cite{nienaber2015detecting}, adding SAM segmentation masks,
geometric feature vectors, VLM-assigned severity classes with confidence
levels, composite risk scores, and natural language warnings for
3,092~assessed potholes in 1,119~scenes. Its labels are generated by
the pipeline; it is intended for developing and comparing methods, not as a
benchmark of physical severity.

\item \textbf{An evaluation against independent references}: physically
measured depth on an RGB-D dataset, a prediction-free three-annotator panel,
segmentation IoU against human-drawn masks, a six-arm incremental prompt
ablation, and a blind human study of the warnings. Several of these results
are negative, and we report them in full.

\end{enumerate}

\subsection{Paper Organization}
\label{sec:organization}

The remainder of this paper is organized as follows.
Section~\ref{sec:related work} reviews related work in pothole detection,
vision-language models, and depth estimation.
Section~\ref{sec:methodology} details the five-stage pipeline.
Section~\ref{sec:experiments} presents experimental validation, including human
studies of severity classification and warning quality.
Section~\ref{sec:results} reports results.
Section~\ref{sec:new_analyses} validates the framework against independent
references and tests its robustness.
Section~\ref{sec:discussion} discusses findings and system limitations.
Section~\ref{sec:cost} analyzes computational costs.
Section~\ref{sec:conclusion} concludes with implications for infrastructure
assessment.

\section{Related Work}
\label{sec:related work}

\subsection{Vision-Language Models for Perception and Depth Reasoning}
\label{sec:vlm_depth}

Vision-Language Models~(VLMs) combine image encoders with large language models
to enable zero-shot transfer and visual question answering. Foundational
architectures such as CLIP~\cite{Radford2021CLIP} and BLIP~\cite{Li2022BLIP}
pioneered contrastive learning on large-scale image-text pairs, while
instruction-tuned successors including LLaVA~\cite{Liu2023LLaVA,Liu2024LLaVA15}
extended these capabilities toward visual reasoning approaching commercial
systems. Commercial models such as GPT
Vision~\cite{OpenAI2023GPT4V} and Google
Gemini~\cite{Google2023Gemini, GoogleAI2025Gemini25} further expanded structured
output generation for API-based deployment. In autonomous driving,
DriveVLM~\cite{TianDriveVLM2024} integrates VLM-based scene description with
hierarchical planning, while PaLM-E~\cite{Driess2023PaLME} generates motion
commands directly from visual and linguistic inputs. Despite these advances,
deploying VLMs in safety-critical structured prediction tasks such as categorical
severity classification remains challenging, with accuracy heavily dependent on the
quality of visual context provided~\cite{Baum2024VLMtransport}.

Accurate depth perception is fundamental to pothole severity assessment, yet
monocular depth estimation from single RGB images is inherently ill-posed: scale
ambiguity prevents metric depth recovery~\cite{Eigen2014, Saxena2005}.
Supervised approaches such as MiDaS~\cite{Ranftl2022MiDaS} and
DPT~\cite{Ranftl2021DPT} achieve strong relative depth accuracy, while
foundation models including Depth Anything~\cite{Yang2024DepthAnythingV1,
Yang2024DepthAnythingV2} and Depth Pro~\cite{Bochkovskiy2024DepthPro} achieve
broad generalization through massive pre-training. However, these networks
optimize for dense pixel-level depth across diverse scenes, not categorical
severity classification within texturally uniform asphalt surfaces where depth
cues are subtle and ambiguous. In our evaluation, foundation depth models do
not resolve pothole depressions on these images
(Sections~\ref{sec:depth_validation} and~\ref{sec:seg_quality}), which motivates
the categorical VLM-based approach adopted in this work.

\subsection{Warning and Guidance Systems for Autonomous Vehicles}
\label{sec:warning_systems}

Traditional Human-Machine Interfaces~(HMIs) for autonomous vehicles rely on
visual icons, auditory tones, and generic warnings that provide limited
contextual information for decision-making~\cite{Louw2020, Lim2023HMI}. Recent
regulatory mandates emphasize multimodal warning systems integrating haptic
feedback, voice alerts, and visual displays~\cite{EU2024ADAS, Singh2023haptic}.

Natural language generation for vehicle hazard warnings has received
comparatively limited attention. Emerging work explores LLMs for real-time
explanations of autonomous vehicle decisions~\cite{Chen2025multimodal,
Meyer2024NLI}, while language-driven driving frameworks demonstrate that LLMs
can interpret commands and translate them into executable
trajectories~\cite{Li2023LaMPilot}. Systems such as
LINGO-1~\cite{Kim2024robustness} combine visual and language modalities to
generate driver-interpretable explanations. However, these systems focus on
explaining autonomous vehicle actions or scene narration rather than generating
geometry-informed hazard warnings with specific maneuver
recommendations tied to road defect characteristics.

\subsection{Positioning of This Work}
\label{sec:positioning}

The reviewed literature reveals three unaddressed gaps. First, no published
system combines VLM-based visual reasoning with deterministic geometric
measurements to resolve monocular scale ambiguity for road defect severity
classification. Second, evaluations of VLMs on structured perception tasks
such as categorical severity rarely use independent references, such as
physical measurements or annotators who have not seen model output. Third, no
systematic evaluation exists comparing vision-only and geometrically grounded
multimodal warnings. Our five-stage modular framework addresses these gaps
through SAM-based segmentation, geometric feature extraction, VLM severity
classification tested against physically measured depth and a prediction-free
human panel, multi-factor risk scoring, and an incremental prompt ablation and
blind human study of the warnings on 1,119~road scenes.

\section{Methodology}
\label{sec:methodology}

\begin{figure*}[t!]
    \centering
    \includegraphics[width=\textwidth]{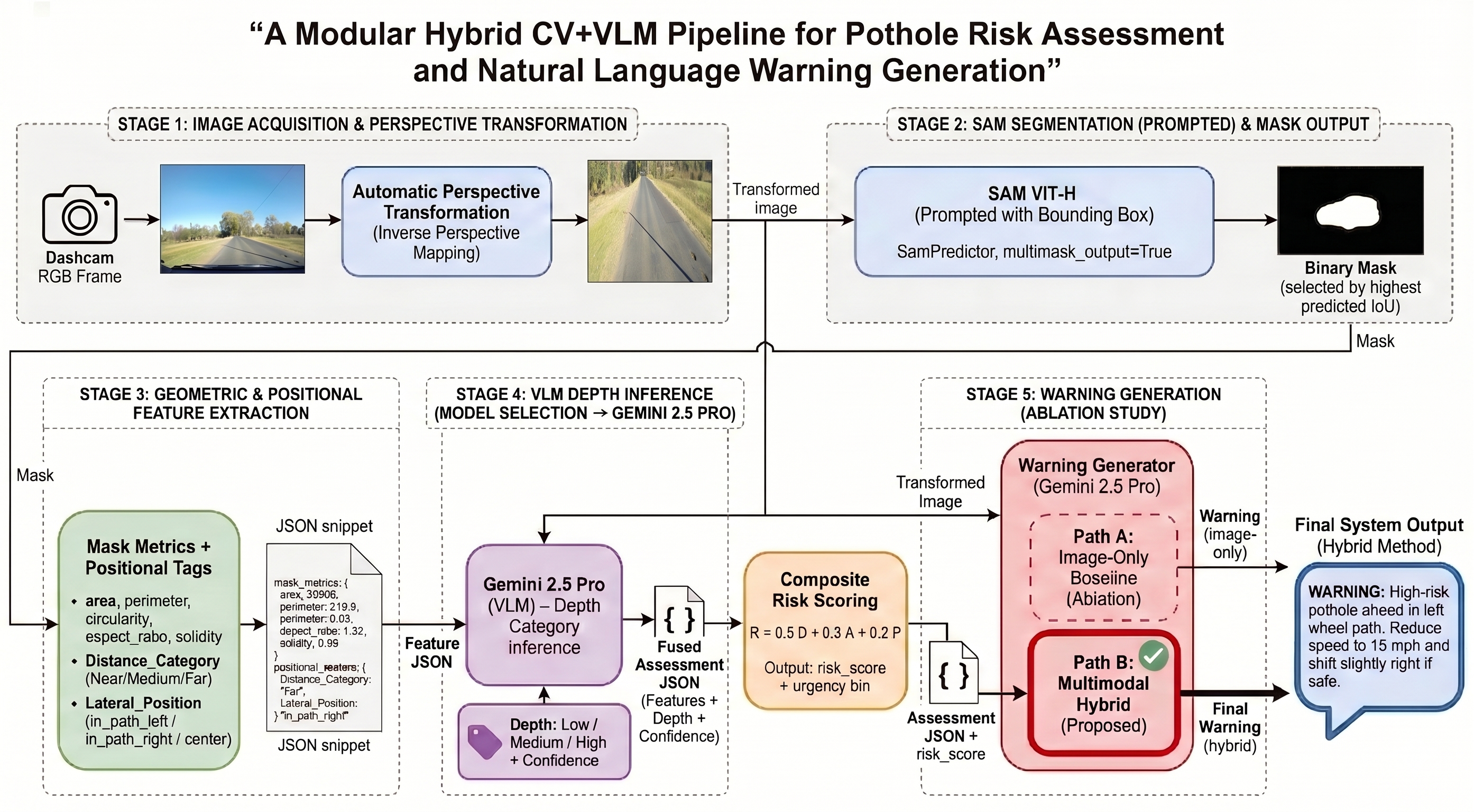}
    \caption{Overview of the five-stage modular pothole hazard assessment
    framework. Stage~1 applies perspective transformation to raw dashcam
    imagery. Stage~2 performs zero-shot pothole segmentation using SAM.
    Stage~3 extracts deterministic geometric features from segmentation masks.
    Stage~4 performs VLM-based severity classification grounded in geometric
    context and computes composite risk scores. Stage~5 generates natural language
    warnings with maneuver recommendations. Structured outputs are passed
    sequentially between stages.}
    \label{fig:pipeline}
\end{figure*}

We propose a five-stage modular framework for offline pothole hazard assessment,
applied to recorded dashcam footage from the public pothole dataset of
Nienaber~et~al.~\cite{nienaber2015detecting}. Each stage operates independently,
consuming the structured output of the preceding stage, which enables isolated
validation and systematic ablation of individual components. Fig.~\ref{fig:pipeline}
illustrates the complete data flow: a dashcam frame undergoes perspective
transformation~(Stage~1), zero-shot pothole segmentation~(Stage~2), deterministic
geometric feature extraction~(Stage~3), VLM-based severity classification with
multi-factor risk scoring~(Stage~4), and finally natural language warning
generation~(Stage~5). The following subsections describe each stage in detail.

\subsection{Stage 1: Perspective Transformation}
\label{sec:stage1}

Raw dashcam imagery is captured from a forward-facing, oblique viewpoint,
which introduces two key challenges for downstream analysis. First, perspective
distortion causes distant potholes to appear disproportionately small relative
to the overall image, resulting in insufficient visual features for reliable
segmentation and geometric measurement. Second, a significant portion of the
image frame is occupied by irrelevant regions such as sky, sidewalks, and roadside
objects that dilute the representation of the region of interest~(ROI) after
rescaling~\cite{AbuRaddaha2024}.

To address these challenges, we apply the automatic perspective transformation
approach developed in our prior work~\cite{AbuRaddaha2024}. The method determines the ROI
automatically from the bounding-box annotations. Specifically, the four corners
of the ROI are derived from the aggregate minimum and maximum $x$- and
$y$-coordinates of the pothole bounding boxes over the training split, expanded
by a proportional offset $\alpha$ to ensure complete coverage of all annotated
regions. These source points are then mapped to the full image corners as target
points, and a $3 \times 3$ homography matrix $\mathbf{M}$ is computed from
the two point sets:
\begin{equation}
    \mathbf{t}' = \mathbf{M} \cdot \mathbf{t},
    \label{eq:homography}
\end{equation}
where $\mathbf{t}$ and $\mathbf{t}'$ are the homogeneous coordinates of a
point in the source and transformed images, respectively. Applying $\mathbf{M}$
warps the image so that the pothole-containing road region expands to fill the
full frame, effectively enlarging distant potholes, sharpening their boundaries,
and reducing the visual weight of irrelevant background regions.

\begin{figure}[t!]
    \centering
    \includegraphics[width=\columnwidth]{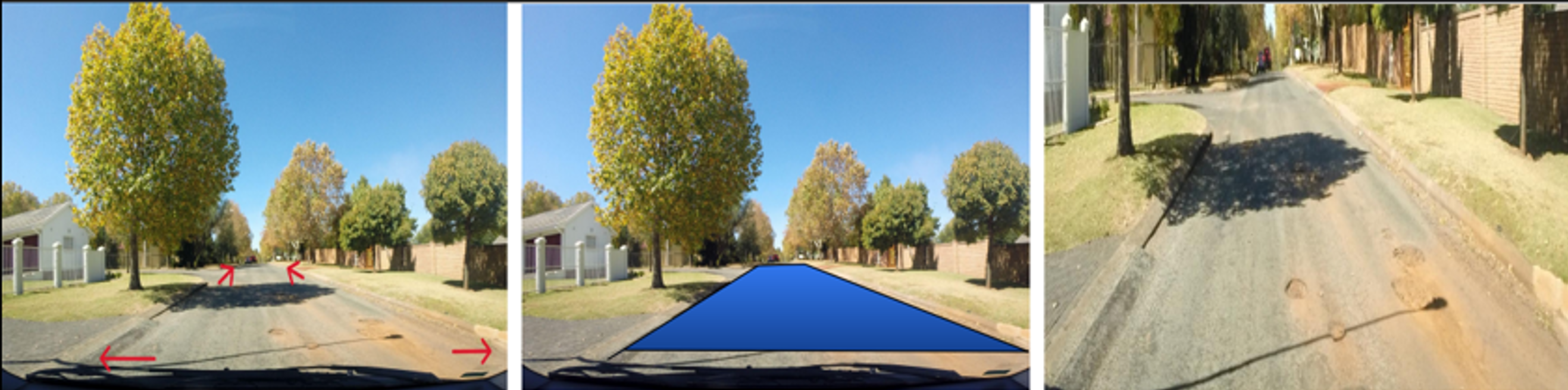}
    \caption{Illustration of the perspective transformation process.
    The transformation maps the pothole-containing road region to fill the
    full frame, reducing foreshortening and improving pothole visibility
    across distances for downstream segmentation and analysis.}
    \label{fig:ipm}
\end{figure}

It is important to note that this transformation is \emph{not} a full
bird's-eye view~(BEV) projection. A complete top-down reprojection would
cause potholes beyond a certain distance to be cropped or distorted,
excluding them from analysis. Instead, the transformation shifts the
viewing perspective to emphasize the road surface while preserving
visibility of both near and far potholes within the same frame. The
matrix $\mathbf{M}$ is computed once, from the aggregate extent of the
bounding-box annotations over the training split, and that single fixed matrix is
then applied to every image in the dataset; it is not recomputed per scene.
Fig.~\ref{fig:ipm} illustrates the perspective transformation process
on a representative dashcam frame, showing how the pothole-containing
road region is expanded to fill the full frame. Full algorithmic
details, including the automated corner-selection procedure and offset
calibration, are provided in~\cite{AbuRaddaha2024}.

\subsection{Stage 2: Zero-Shot Pothole Segmentation}
\label{sec:stage2}

Following perspective transformation, we apply Meta's Segment Anything
Model~(SAM)~\cite{Kirillov2023} with the ViT-H backbone for pothole boundary
delineation on the transformed imagery. SAM's foundation
model architecture trained on over 1.1~billion masks spanning diverse visual
domains, enables zero-shot generalization to road defect segmentation without
task-specific fine-tuning or retraining~\cite{Zhang2023SAMsurvey}. This is
particularly important for deployment robustness, as supervised segmentation
networks trained on one pavement type or geographic region often fail to
generalize to different road materials or damage patterns~\cite{Ke2023HQ}.

\begin{figure}[t!]
    \centering
    \includegraphics[width=\columnwidth]{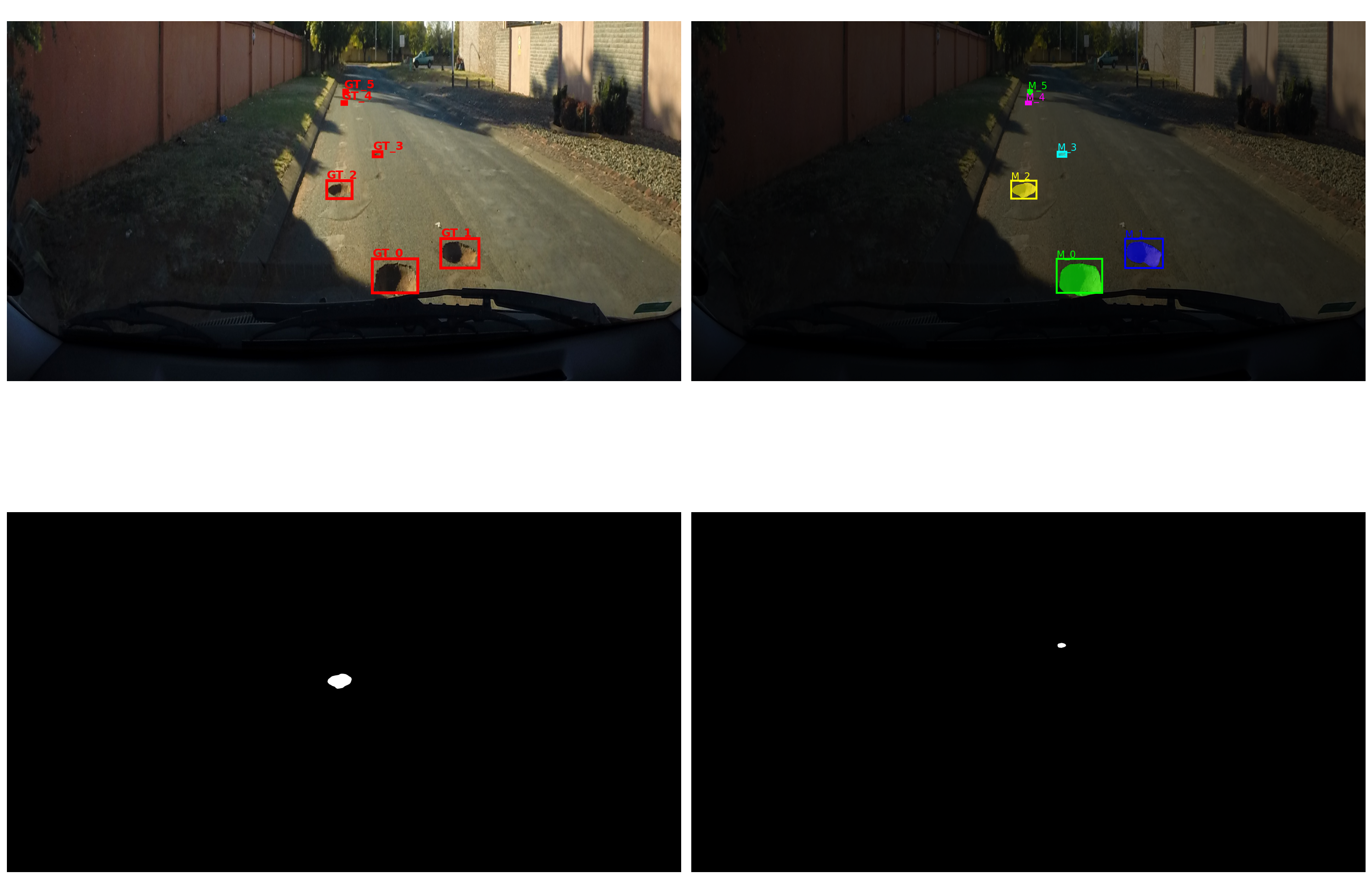}
    \caption{Example SAM segmentation output on a perspective-transformed
    road scene. Left: transformed image with ground truth bounding box
    prompt. Right: Generated masks in different colors, and the bottom
    row shows two mask samples where the left one has a score of
    0.970, and the right one has a score of 0.891.}
    \label{fig:sam_output}
\end{figure}

We operate SAM in prompted mode using \texttt{SamPredictor}, providing
axis-aligned bounding boxes $\mathbf{B} = (x_1, y_1, x_2, y_2)$ derived
from the ground truth annotations of the dataset as spatial prompts. For
each bounding box, SAM generates three mask candidates with
\texttt{multimask\_output=True}, each accompanied by a predicted
Intersection-over-Union~(IoU) confidence score. We select the mask
$\hat{M}$ with the highest predicted IoU:
\begin{equation}
    \hat{M} = \arg\max_{M_i,\, i \in \{1,2,3\}} \text{IoU}_{\text{pred}}(M_i)
    \label{eq:sam_select}
\end{equation}
This selection uses SAM's internal quality estimate to choose among the
candidates, which helps in ambiguous cases where pothole edges transition
gradually into intact pavement. The predicted IoU is a confidence estimate, not
a measurement: against human-drawn masks the selected masks reach a mean IoU of
0.780, and the predicted score overestimates it by 0.17 on average
(Section~\ref{sec:seg_quality}). The output of Stage~2 for each pothole is
a binary segmentation mask $M \in \{0,1\}^{H \times W}$, where foreground
pixels ($M(x,y) = 1$) represent the pothole region, paired with its
ground truth bounding box~$\mathbf{B}$ and predicted IoU score
$s_{\text{IoU}} \in [0,1]$.
Fig.~\ref{fig:sam_output} shows a canonical example of SAM segmentation
applied to a perspective-transformed road scene, including the bounding
box prompt, the resulting multi-mask candidates, and two representative
mask samples with their predicted IoU scores.

\subsection{Stage 3: Deterministic Geometric Feature Extraction}
\label{sec:stage3}

Geometric feature extraction transforms the binary SAM mask from Stage~2 into
a structured six-dimensional feature vector that characterizes each pothole's
shape, size, and spatial relationship to the vehicle's trajectory. These
deterministic measurements are computed via classical computer vision operators
rather than learned embeddings, ensuring interpretability and robustness across
diverse road conditions. The feature vector is serialized as a JSON object and
passed as structured context to the VLM in Stage~4. These quantities are measured
in the perspective-transformed image domain and are therefore
\emph{image-geometry-informed} rather than metric: they carry no physical scale
and do not by themselves resolve monocular depth ambiguity. What they provide is
relative morphological and spatial context that constrains the model's
interpretation and is intended to make its severity judgments more
consistent~\cite{SpatialRGPT2024, Baum2024VLMtransport}.

\subsubsection{Shape Descriptors}
We extract the largest external contour $C$ from the binary mask, which
isolates the pothole boundary while discarding interior holes or texture
artifacts. Five shape metrics characterize defect morphology:

\begin{enumerate}

    \item \textbf{Area} ($A$): total foreground pixel count,
    $A = \sum_{(x,y)} M(x,y)$, providing a relative measure of pothole extent
    in the perspective-transformed frame.

    \item \textbf{Perimeter} ($P$): arc length of the external contour $C$,
    computed via \texttt{arcLength}.

    \item \textbf{Circularity} ($\mathcal{C}$): shape compactness defined as
    $\mathcal{C} = 4\pi A / P^{2}$, where $\mathcal{C} = 1$ indicates a
    perfect circle and values approaching zero correspond to elongated or
    irregular forms such as linear cracks.

    \item \textbf{Aspect Ratio} ($\mathrm{AR}$): ratio of the bounding box
    width to height, $\mathrm{AR} = W_{\mathbf{B}} / H_{\mathbf{B}}$,
    quantifying shape elongation.

    \item \textbf{Solidity} ($S$): ratio of mask area to convex hull area,
    $S = A / A_{\text{hull}}$, measuring boundary irregularity and
    concavity. A solidity value near 1.0 indicates a compact, convex shape,
    while lower values suggest fragmented or concave boundaries characteristic
    of compound potholes or interconnected defects.

\end{enumerate}

\subsubsection{Spatial Positioning and Trajectory Intersection}
The sixth feature characterizes each pothole's position relative to the
estimated vehicle trajectory, combining both longitudinal distance estimation
and lateral zone classification.

\textbf{Longitudinal Distance Category.} Following the distance classification
methodology of our prior work~\cite{AbuRaddaha2024}, the perspective-transformed
image is divided into three horizontal bands based on the $y$-coordinate
$y_2$ of the bottom edge of the pothole bounding box. For the automatically
transformed images (1,965\,px high), potholes with $y_2 \leq 449$ are classified
as \emph{far}, those with $450 \leq y_2 \leq 514$ as \emph{medium}, and those
with $y_2 \geq 515$ as \emph{near}; this rule reproduces the released labels for
every pothole. The complete derivation details are provided in~\cite{AbuRaddaha2024}.
These bands are ordinal proximity categories defined by image rows under a
fixed camera geometry; they are not calibrated metric distances.

\textbf{Lateral Position Category.} The horizontal extent of the pothole
bounding box, defined by its left x-coordinate $x_{\min}$ and right
x-coordinate $x_{\max}$ in the perspective-transformed image frame, is used
to assign the pothole to one of four lateral zones relative to the estimated
vehicle wheel paths. The zone widths are motivated by vehicle lateral-position
data: on a standard 12-ft~(3.65\,m) lane, each wheel track occupies
approximately 20\% of the lane width, with both tracks together covering
roughly 40--50\% of the total lane~\cite{HilgertSAE2008,
FHWA2013DistressManual}. Accordingly, the image width is divided into four
zones: \emph{in-path left} and \emph{in-path right} (each approximately 20\%
of the image width), representing the left and right wheel track regions where
direct vehicle contact is most probable, and \emph{far-left} and
\emph{far-right} (each approximately 30\% of the image width) at the outer
edges of the frame, where collision risk is substantially lower. This zoning
assumes an approximately straight road, a fixed forward-facing camera, and a
vehicle traveling near the center of its lane; it does not model camera pose,
lane boundaries, road curvature, vehicle width, or the vehicle's lateral
position, and is therefore an approximation (Section~\ref{sec:limitations}).
A pothole is assigned to the zone containing the left edge $x_{\min}$ of its
bounding box, the rule used to generate the released labels and implemented in
the released code; assigning by the box center instead would change the zone of
8.3\% of potholes, all at zone boundaries. Potholes within
either in-path zone are treated as high-priority hazards in the downstream
risk scoring stage.

The complete six-dimensional feature vector
$\mathbf{f} = [A,\; P,\; \mathcal{C},\; \mathrm{AR},\; S,\; \text{pos}]$,
where $\text{pos}$ encodes both longitudinal distance category and lateral
zone, is summarized in Table~\ref{tab:features}.

\begin{table}[!t]
\caption{Six-Dimensional Geometric Feature Vector}
\label{tab:features}
\centering
\setlength{\tabcolsep}{4pt}
\renewcommand{\arraystretch}{1.2}
\begin{tabular}{clp{4.6cm}}
\hline
\textbf{\#} & \textbf{Feature} & \textbf{Description} \\
\hline
1 & Area ($A$)                        & Foreground pixel count of SAM mask \\
2 & Perimeter ($P$)                   & Arc length of external contour \\
3 & Circularity ($\mathcal{C}$)       & Shape compactness: $4\pi A / P^{2}$ \\
4 & Aspect Ratio ($\mathrm{AR}$)      & Bounding box width-to-height ratio \\
5 & Solidity ($S$)                    & Mask area / convex hull area \\
6 & Position (\text{pos})             & Joint encoding of longitudinal distance
                                        (Near / Medium / Far) and lateral zone
                                        (Far-Left / In-Path Left /
                                        In-Path Right / Far-Right) \\
\hline
\end{tabular}
\end{table}

\subsection{Stage 4: VLM-Based Severity Classification and Risk Scoring}
\label{sec:stage4}

Stage~4 performs two tightly coupled operations: categorical severity
classification using a vision-language model and composite risk scoring that
combines severity with geometric and spatial features from Stage~3.

\subsubsection{Severity Classification}

Stage~4 assigns each pothole one of three severity classes, \emph{low},
\emph{medium}, or \emph{high}. The classes are named after the depth bands
$<$2\,cm, 2--5\,cm, and $>$5\,cm, following the depth-based pothole severity
convention of pavement distress manuals, which grade potholes by depth (for
example, low $<$25\,mm, moderate 25--50\,mm, and high $>$50\,mm in the FHWA
manual~\cite{FHWA2013DistressManual}; see also~\cite{OntarioReg2002}).
Because the model sees only an RGB image, the output is a perceived
visual-severity class, not a depth measurement;
Section~\ref{sec:metric_validation} tests it against physically measured depth.
For brevity we refer to this step as depth inference. We employ Google
Gemini~2.5~Pro as the engine, selected through a comparative evaluation of
three frontier vision-language models against a three-annotator human panel;
the validation methodology and results are presented in
Section~\ref{sec:experiments}. Gemini~2.5~Pro was chosen because its categorical severity agreement with the
human consensus was the highest of the models compared, while retaining the
structured JSON output capabilities required for downstream risk scoring.

\textbf{Adaptive Image Input Strategy.}
Since cropped pothole images from distant or small defects often lack
sufficient visual detail for reliable texture analysis, we implement an
adaptive dual-image strategy. For each pothole, the system evaluates two
conditions: whether the pothole is classified as \emph{far} by the distance
category from Stage~3, or whether the minimum dimension of the cropped
region falls below 100\,pixels. If either condition holds, the VLM receives
both the full perspective-transformed scene, providing spatial context and
scene-level lighting reference, alongside the cropped pothole patch. For
near and medium potholes whose crops contain sufficient visual detail, only
the cropped region is transmitted to reduce API token consumption. This
per-pothole inference design ensures that depth classification is performed
independently for each defect, preventing scene-level visual bias from
influencing individual severity assessments.

\textbf{Multimodal Input Fusion.}
In addition to the image input, every API call includes the geometric feature
vector from Stage~3 serialized as a structured JSON object, providing the VLM
with explicit shape and positional context. This fusion is motivated by the
known limitation of monocular depth inference. VLMs operating on image data
alone lack real-world scale references and tend toward conservative
over-prediction under visual ambiguity~\cite{Baum2024VLMtransport,
SpatialRGPT2024}. The geometric features, particularly mask area, circularity,
solidity, aspect ratio, and distance category, serve as relative scale proxies
that constrain the VLM's visual interpretation, with the aim of reducing
shadow-induced overestimation in shallow potholes and of weighting large,
irregular masks toward higher severity categories.

\textbf{Prompting Strategy.}
The prompt supplied to Gemini~2.5~Pro is structured around four components:
(i)~a scientific system context preamble, included to prevent false
rejections of road damage imagery by the model's content safety filters;
(ii)~the injected geometric feature values from Stage~3 with operational
definitions for each metric; (iii)~explicit depth category definitions
grounded in the physical thresholds of Table~\ref{tab:depth_categories};
and (iv)~a visual reasoning directive specifying the image cues the model
should prioritize, including shadow depth, edge sharpness, surface texture, and
interior darkness to maintain classification consistency across varying
lighting conditions. The complete prompt template is provided in
Appendix~\ref{appendix:prompts}.

The API is configured to return schema-constrained JSON responses
containing three mandatory fields. The \texttt{depth\_category} field
$\in$~\{\emph{low}, \emph{medium}, \emph{high}\} records the predicted
severity. A self-reported \texttt{confidence} level
$\in$~\{\emph{low}, \emph{medium}, \emph{high}\} reflects the model's
certainty given the available visual and geometric evidence, and a
\texttt{reasoning} string cites the specific visual cues used. The confidence
field is produced by the VLM itself based on the clarity of the image and the
consistency between visual appearance and the provided geometric features; it
is retained as metadata for downstream interpretability. It was also
displayed, together with the models' predicted classes, to the annotators of
the first depth-validation panel (Section~\ref{sec:depth_validation}), which
limits that panel's independence; the second panel saw no model output
(Section~\ref{sec:v1}). Generation temperature is set to 0.1 to reduce
run-to-run variation; a low temperature does not make a hosted model
deterministic, and Section~\ref{sec:variability} measures the remaining
variation. Parsing failures and safety-filter blocks, which occurred in under
5\% of calls during pilot testing, default to a \emph{medium} class with
\emph{low} confidence and are logged for offline review; in the full
3,092-pothole run no call was blocked or failed to parse, so this default was
never applied.

\begin{table}[!t]
\caption{Severity Category Operational Definitions}
\label{tab:depth_categories}
\centering
\setlength{\tabcolsep}{4pt}
\renewcommand{\arraystretch}{1.2}
\begin{tabular}{p{1.1cm} p{1.4cm} p{5.0cm}}
\hline
\textbf{Category} & \textbf{Nominal Depth} & \textbf{Visual Characteristics} \\
\hline
Low    & $<$2\,cm   & Shallow surface defect; minimal shadow; slight texture variation; no visible cavity floor \\
Medium & 2--5\,cm   & Moderate depression; visible shadow; discernible boundary definition; partial interior visibility \\
High   & $>$5\,cm   & Deep cavity; pronounced shadow; rough or irregular interior surface; potential structural undermining \\
\hline
\end{tabular}
\end{table}

\subsubsection{Composite Risk Scoring}

The composite risk score aggregates depth inference with the geometric and
spatial features from Stage~3 into a single numeric hazard index per pothole,
enabling prioritization of multiple defects within the same scene and
providing a quantitative input for the natural-language warning generator in
Stage~5. Rather than relying on any single cue, the score fuses depth
severity, spatial extent, and positional proximity into a weighted linear
combination intended to reflect the relative contribution of each factor to
driving hazard.

Formally, the composite risk score $R$ for a given pothole is defined as
\begin{equation}
    R \;=\; w_d \, S_d \;+\; w_a \, S_a \;+\; w_p \, S_p,
    \label{eq:risk_score}
\end{equation}
where $S_d$ is the depth severity score, $S_a$ is the area severity score,
$S_p$ is the spatial position score, and $w_d = 0.5$, $w_a = 0.3$,
$w_p = 0.2$ are the corresponding weights. The weights are design choices,
not fitted parameters: no outcome data, such as recorded vehicle damage, exist
with which to fit them. Severity receives the largest weight because pavement
distress manuals grade pothole severity primarily by
depth~\cite{FHWA2013DistressManual, OntarioReg2002}.
Section~\ref{sec:risk_sensitivity} reports how the outputs change when the
weights and category boundaries are perturbed.

The severity score $S_d$ maps the class to a number:
\emph{low}~$\rightarrow 20$, \emph{medium}~$\rightarrow 60$,
\emph{high}~$\rightarrow 100$.

The area score $S_a$ is a fixed logarithmic function of the mask area $A$ in
pixels, $S_a = 100\cdot\mathrm{clip}\big((\log_{10}A - 2)/2.5,\,0,\,1\big)$,
which rises from 0 at 100~pixels to 100 at about 31{,}600~pixels. The logarithm
keeps a few very large potholes from dominating the score, and because the
constants are fixed in advance, no quantity estimated from the evaluation data
enters the score.

The position score $S_p$ depends on the lateral zone from Stage~3: 100 for
potholes in either wheel-path zone and 70 for potholes in the outer zones (a
value of 30 is reserved for positions outside the four zones, which does not
occur in ENPAD). Longitudinal distance does not enter $R$; it is supplied to
the warning generator in Stage~5 and used by the escalation rules of
Table~\ref{tab:urgency_rubric}. The complete weight and threshold
specifications are summarized in Table~\ref{tab:risk_weights}.

\begin{table}[!t]
\caption{Composite Risk Score Components, Weights, and Thresholds}
\label{tab:risk_weights}
\centering
\setlength{\tabcolsep}{4pt}
\renewcommand{\arraystretch}{1.15}
\begin{tabular}{p{1.8cm} c p{4.5cm}}
\hline
\textbf{Component} & \textbf{Weight} & \textbf{Score Mapping} \\
\hline
Depth ($S_d$)    & 0.5 & Low$\to$20,\; Medium$\to$60,\; High$\to$100 \\
Area ($S_a$)     & 0.3 & $100\cdot\mathrm{clip}\!\left(\frac{\log_{10}A - 2}{2.5},0,1\right)$ \\
Position ($S_p$) & 0.2 & In-path$\to$100,\; In-lane (outside wheel path)$\to$70,\; Other$\to$30 \\
\hline
\multicolumn{3}{l}{\footnotesize Risk categories: Low ($R < 40$), Medium ($40 \leq R < 60$),} \\
\multicolumn{3}{l}{\footnotesize High ($60 \leq R < 80$), Critical ($R \geq 80$).} \\
\hline
\end{tabular}
\end{table}

The resulting score $R$ is mapped to four ordinal risk categories with
fixed boundaries: \emph{Low} ($R < 40$), \emph{Medium}
($40 \leq R < 60$), \emph{High} ($60 \leq R < 80$), and \emph{Critical}
($R \geq 80$). These thresholds partition the 0--100 score range into
operationally meaningful tiers that correspond to increasing urgency of driver
response, from routine awareness to immediate evasive action. At the scene
level, the overall risk is determined by the maximum individual pothole risk
across all detected defects, ensuring that the most hazardous pothole in a
multi-defect scene governs the scene-level warning. When the trigger
conditions of Table~\ref{tab:urgency_rubric} are applied, the numeric band is
primary and the trigger conditions act only to escalate a pothole to a higher
level, never to lower it; the two agree for 92.6\% of potholes. The per-pothole risk
score, risk category, and all upstream metadata are packaged as a structured
JSON record and forwarded to Stage~5 for natural language generation.

\subsection{Stage 5: Natural Language Warning Generation}
\label{sec:stage5}

The final stage translates the structured risk assessment into
context-sensitive natural language warnings that communicate hazard severity,
recommended driving actions, and supporting rationale. While the preceding
stages produce quantitative outputs suitable for automated decision systems,
effective communication to human operators requires prose that conveys urgency
proportionally and provides actionable guidance without overwhelming the
reader with raw numeric data.

We employ Google Gemini~2.5~Pro as the generation engine for this stage,
the same model used for depth inference in Stage~4. Gemini~2.5~Pro was
selected for its strong language generation quality, producing specific and
contextually grounded professional warnings that reference the quantitative
assessment data. The model's ability to integrate both visual scene context
and structured numeric inputs makes it well-suited for generating warnings
that go beyond generic templates.

The generation prompt follows a multimodal hybrid design that combines visual
and structured numeric inputs. Each API call receives the
perspective-transformed scene image alongside per-pothole structured data
including depth category, VLM confidence, geometric features from Stage~3,
composite risk score and category from Stage~4, and spatial position
information. This dual-channel input enables the model to ground its warnings
in both the visual scene context and the precise quantitative assessment,
producing warnings that reference specific spatial relationships visible in
the image rather than generating generic templated text. An image-only
baseline variant that omits the structured data is evaluated for ablation
comparison in Section~\ref{sec:experiments}. The complete prompt template,
including the domain knowledge reference and worked examples, is provided in
Appendix~\ref{appendix:prompts}.

The prompt incorporates several design elements intended to make the output
consistent. A domain knowledge reference section provides the model
with the operational definitions of depth categories, distance categories with
associated approximate reaction times, lateral position semantics, and risk
score interpretation ranges. A decision-making rubric maps risk score ranges
to expected urgency levels, as summarized in Table~\ref{tab:urgency_rubric},
intended to keep the model's urgency consistent with the quantitative
assessment. Four worked examples spanning critical, high, medium,
and low risk scenarios establish the expected tone, specificity, and length
for each urgency tier. The incremental ablation of
Section~\ref{sec:prompt_ablation} shows that removing these examples changes
the rubric consistency of the generated urgency by only 0.3 percentage points;
their role is mainly to set tone and length.

\begin{table}[!t]
\caption{Decision Rubric Mapping Risk Scores to Urgency Levels (the numeric band
is primary; trigger conditions can only escalate)}
\label{tab:urgency_rubric}
\centering
\setlength{\tabcolsep}{4pt}
\renewcommand{\arraystretch}{1.2}
\begin{tabular}{p{1.3cm} p{1.5cm} p{4.5cm}}
\hline
\textbf{Urgency} & \textbf{Risk Score} & \textbf{Trigger Conditions} \\
\hline
Critical & $R \geq 80$ & High depth + Near distance + In wheel path, or composite risk $\geq 80$ \\
High     & $60$--$79$  & Medium/high depth + Near distance, or composite risk 60--79 \\
Medium   & $40$--$59$  & Low/medium depth + Near, or high depth + Far, or composite risk 40--59 \\
Low      & $R < 40$    & Low depth + Medium/far distance, or composite risk $< 40$ \\
\hline
\end{tabular}
\end{table}

The model is configured to return a six-field JSON response for each scene.
The \texttt{warning\_message} field contains a one-to-two sentence
professional warning summarizing the hazard. The
\texttt{maneuver\_suggestion} field provides a specific driving action such
as lane adjustment, speed reduction, or heightened alertness appropriate to
the risk level. The \texttt{urgency\_level} field
$\in$~\{\emph{critical}, \emph{high}, \emph{medium}, \emph{low}\} enables
downstream systems to select visual or auditory alert modalities accordingly.
The \texttt{reasoning} field provides an explanation citing the specific
depth, area, and position values that informed the assessment, supporting
transparency and trust in the system's recommendations. The
\texttt{affected\_lanes} field identifies which lanes contain hazards, and
the \texttt{recommended\_speed\_reduction} field specifies a proportional
speed adjustment. The hybrid warnings were generated with the API's default
sampling settings. Parsing failures and safety-filter blocks default to
a \emph{medium} urgency level with a generic advisory warning and are logged
for offline review; in the 1,119-scene run all responses parsed, so this
default was never applied.

\section{Experiments}
\label{sec:experiments}

This section describes the dataset, the validation protocol for VLM-based
severity classification (Stage~4), and the ablation study design for warning
generation (Stage~5). Quantitative results and statistical analysis are
reported in Section~\ref{sec:results}.

\subsection{Dataset}
\label{sec:dataset}

All experiments are conducted on the publicly available pothole dataset
of Nienaber~et~al.~\cite{nienaber2015detecting}, which contains 1,119
dashcam scenes captured on South African road networks with 3,097 manually
annotated potholes, of which 3,092 are assessed (three boxes lie outside the
transformed frame and two were not processed). We extend this dataset with automatically computed
annotations produced by the five-stage pipeline described in
Section~\ref{sec:methodology}, yielding the \textbf{Extended Nienaber
Pothole Assessment Dataset~(ENPAD)}. ENPAD is an \emph{annotation resource}
whose severity labels are generated by the pipeline described in
Section~\ref{sec:methodology} and compared with human judgment only on the
102 potholes of the two human panels (Sections~\ref{sec:human_validation}
and~\ref{sec:v1}); it is not an
independently verified benchmark of physical pothole severity, and we do not
propose it as one. Its intended use is as a structured corpus for developing
and comparing severity-assessment and warning-generation methods. For each
pothole in ENPAD, the
following structured attributes are stored:

\begin{itemize}[noitemsep, topsep=3pt]
    \item Perspective-transformed imagery (Stage~1),
    \item SAM segmentation mask and predicted IoU score (Stage~2),
    \item six-dimensional geometric feature vector
          $\mathbf{f} = [A,\; P,\; \mathcal{C},\; \mathrm{AR},\; S,\;
          \text{pos}]$ (Stage~3),
    \item VLM-inferred depth category with associated confidence level
          and composite risk score with severity label (Stage~4),
    \item natural language warning message with urgency level and
          maneuver suggestion (Stage~5).
\end{itemize}

Table~\ref{tab:dataset_stats} summarizes the distribution of the 3,092
potholes across the categorical attributes that govern risk scoring.
The distribution is strongly skewed. The \emph{low} severity class accounts
for 74.9\% of potholes and the \emph{high} class for only 5.6\%, which is
consistent with the conservative labeling behavior we quantify against
physically measured depth in Section~\ref{sec:metric_validation}: the model
under-calls the deepest defects, and that bias is visible in the corpus
statistics as well as in the controlled validation. Longitudinal distance is
likewise skewed rather than balanced, with 82.9\% of potholes in the
\emph{Near} band---an expected consequence of the perspective transformation,
which expands the near field and compresses the far field. Lateral position
shows a pronounced concentration within the two in-path zones (81.3\% of
potholes), consistent with the expectation that wheel loading concentrates
damage in the wheel paths.

\begin{table}[!t]
\caption{ENPAD Dataset Statistics}
\label{tab:dataset_stats}
\centering
\setlength{\tabcolsep}{5pt}
\renewcommand{\arraystretch}{1.15}
\begin{tabular}{l l r r}
\hline
\textbf{Attribute} & \textbf{Category} & \textbf{Count} & \textbf{\%} \\
\hline
\multirow{3}{*}{Depth}
    & Low            & 2,316  & 74.9 \\
    & Medium         & 603    & 19.5 \\
    & High           & 173    & 5.6 \\
\hline
\multirow{3}{*}{Distance}
    & Near           & 2,562  & 82.9 \\
    & Medium         & 158    & 5.1 \\
    & Far            & 372    & 12.0 \\
\hline
\multirow{4}{*}{Lateral Position}
    & Far-Left       & 205    & 6.6 \\
    & In-Path Left   & 960    & 31.0 \\
    & In-Path Right  & 1,553  & 50.2 \\
    & Far-Right      & 374    & 12.1 \\
\hline
\multicolumn{2}{l}{\textbf{Total Scenes}}   & \multicolumn{2}{r}{1,119} \\
\multicolumn{2}{l}{\textbf{Total Potholes}} & \multicolumn{2}{r}{3,092} \\
\hline
\end{tabular}
\end{table}

\begin{table}[!t]
\caption{Per-class precision/recall against the first panel's consensus ($n=38$). The first three rows are the validation run; the last row scores the released full-dataset labels. Confusion matrices (rows: consensus low/medium/high; columns: model): validation run $[13, 0, 0; 4, 8, 3; 3, 3, 4]$, released labels $[12, 1, 0; 5, 6, 4; 4, 3, 3]$.}
\label{tab:vlm_perclass}
\centering
\setlength{\tabcolsep}{5pt}
\resizebox{\columnwidth}{!}{%
\begin{tabular}{l c c c}
\hline
\textbf{Model} & \textbf{Low P/R} & \textbf{Medium P/R} & \textbf{High P/R} \\
\hline
Gemini 2.5 Pro & 0.65/1.00 & 0.73/0.53 & 0.57/0.40 \\
Gemini 2.5 Flash & 0.69/0.69 & 0.47/0.60 & 0.17/0.10 \\
GPT-5 & 0.42/1.00 & 0.86/0.40 & 0.00/0.00 \\
\hline
Gemini 2.5 Pro (released) & 0.57/0.92 & 0.60/0.40 & 0.43/0.30 \\
\hline
\end{tabular}}
\end{table}

\subsection{VLM Severity Classification Validation (Stage~4)}
\label{sec:depth_validation}

\subsubsection{Foundation Depth Model Ablation}

Before adopting VLM-based categorical depth inference, we investigated
whether established monocular depth estimation models could directly
extract pothole depth information from dashcam imagery. We evaluated three
state-of-the-art foundation models: Depth Anything
v1~\cite{Yang2024DepthAnythingV1}, Depth Anything v2~\cite{Yang2024DepthAnythingV2},
and Depth Pro~\cite{Bochkovskiy2024DepthPro}. These models produce dense
per-pixel depth maps from single RGB images and have demonstrated strong
performance on standard depth estimation benchmarks.

\begin{figure*}[t!]
    \centering
    \begin{minipage}{0.32\textwidth}
        \centering
        \includegraphics[width=\textwidth]{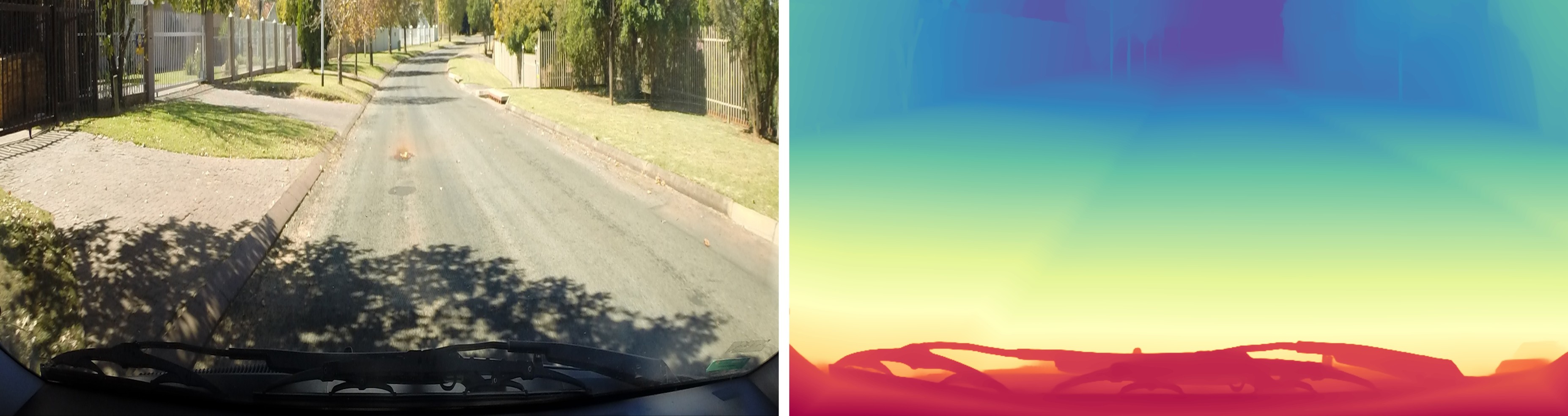}
        \centerline{\small (a) Depth Anything v1}
    \end{minipage}\hfill
    \begin{minipage}{0.32\textwidth}
        \centering
        \includegraphics[width=\textwidth]{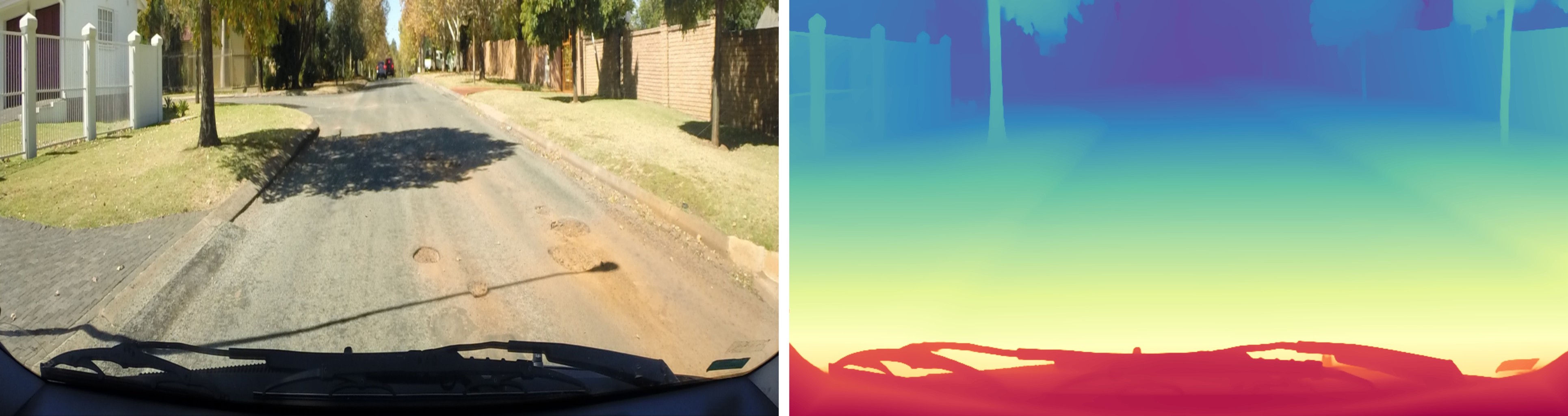}
        \centerline{\small (b) Depth Anything v2}
    \end{minipage}\hfill
    \begin{minipage}{0.32\textwidth}
        \centering
        \includegraphics[width=\textwidth]{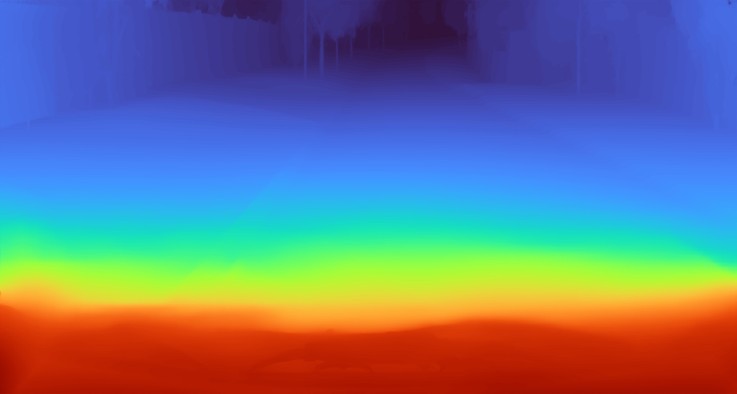}
        \centerline{\small (c) Depth Pro}
    \end{minipage}
    \caption{Depth maps produced by foundation monocular depth estimation
    models on pothole scenes. All three models smooth over pothole
    depressions, treating them as flat road surface. The potholes are
    effectively invisible in the predicted depth maps, illustrating that, on
    these examples, the models do not capture the localized depth
    discontinuities required for pothole severity assessment.}
    \label{fig:depth_ablation}
\end{figure*}

On our images, all three models performed poorly on pothole depth recovery.
As illustrated in Fig.~\ref{fig:depth_ablation}, the predicted depth maps
smooth over pothole depressions, treating them as flat road
surface rather than as localized cavities. The models effectively act as
if the potholes are not present, producing depth maps where the road
surface appears uniformly planar. This behavior is expected: these models
are trained on large-scale datasets where road surfaces are predominantly
flat, and small localized depressions at the centimeter scale fall below
the spatial resolution and training distribution of the depth estimators.
Potholes are effectively treated as noise rather than as meaningful depth
discontinuities. Fig.~\ref{fig:depth_ablation} is a qualitative illustration;
Section~\ref{sec:seg_quality} quantifies the effect for Depth Anything~v2
against sensor-measured depth, whereas Depth Anything~v1 and Depth Pro were
assessed only qualitatively.

This ablation finding motivates our adoption of VLM-based categorical
depth inference, which leverages visual reasoning about shadow patterns,
edge characteristics, and surface texture to classify depth severity
rather than attempting to recover metric depth values from a fundamentally
ill-posed monocular estimation problem.

\subsubsection{Human-Referenced Validation}
\label{sec:human_validation}

Monocular pothole depth estimation from dashcam imagery is inherently
subjective, and no depth ground truth exists in the source
dataset~\cite{nienaber2015detecting}. We therefore compare three frontier
vision-language models against a human panel on a common sample and select
the model with the highest agreement for full-dataset processing. Two
limitations of this first panel are addressed later in the paper: its
annotation sheet displayed the models' predictions, and it is small.
Sections~\ref{sec:metric_validation} and~\ref{sec:v1} add a physical-depth
reference and a larger, prediction-free panel.

\subsubsection{Human Annotation Study}

Three annotators (two computer science graduate students and one
professional driver with over 15~years of experience) classified 42 randomly
sampled potholes into the depth categories of
Table~\ref{tab:depth_categories} (\emph{low}, \emph{medium},
\emph{high}). Annotators were presented with the same
perspective-transformed images and cropped pothole patches supplied to
the VLMs, along with the operational depth definitions and visual cue
guidelines reproduced in Appendix~\ref{appendix:depth_prompt}. No
geometric feature data were provided to human annotators. The annotation sheet
did, however, list each model's predicted class and Gemini~2.5~Flash's
confidence, and the instructions asked annotators to consider all available
information. The labels of this panel are therefore not independent of the
models; Sections~\ref{sec:v1} and~\ref{sec:depth_limitations} quantify the
consequence.

Inter-annotator agreement was measured by Cohen's kappa
($\kappa$)~\cite{Cohen1960kappa} for each annotator pair. Cohen's kappa
quantifies the degree of agreement between two raters beyond what would
be expected by chance alone:
\begin{equation}
    \kappa = \frac{p_o - p_e}{1 - p_e},
    \label{eq:cohens_kappa}
\end{equation}
where $p_o$ is the observed proportion of agreement (the fraction of
items on which both raters assign the same category) and $p_e$ is the
expected proportion of agreement under random assignment, computed from
the marginal distributions of the two raters'
labels~\cite{Cohen1960kappa}. Values of $\kappa = 1$ indicate perfect
agreement, $\kappa = 0$ indicates agreement no better than chance, and
negative values indicate systematic disagreement. Following the widely
adopted interpretation scale of Landis and
Koch~\cite{Landis1977kappa}: $\kappa < 0.20$ is slight, $0.21$--$0.40$
is fair, $0.41$--$0.60$ is moderate, $0.61$--$0.80$ is substantial,
and $0.81$--$1.00$ is almost perfect agreement.

The three pairwise scores and raw agreement percentages are reported in
Table~\ref{tab:human_agreement}. The average pairwise kappa of
$\bar{\kappa} = 0.453$ falls within the \emph{moderate agreement}
band~\cite{Landis1977kappa}, indicating that severity judgment from a single
image is ambiguous for human observers as well (the later prediction-free panel
of Section~\ref{sec:v1} agreed more closely). Of the 42 potholes, 21 (50.0\%)
received unanimous agreement across all three annotators, 17 (40.5\%) reached
majority consensus (two of three agree), and 4 (9.5\%) produced a three-way
split with no majority label, leaving 38 consensus items. For subsequent VLM evaluation, the human
consensus label is defined as the majority category when at least two
annotators agree; three-way splits are excluded from the consensus
comparison.

\begin{table}[!t]
\caption{Human Inter-Annotator Agreement on Depth Classification ($n=42$)}
\label{tab:human_agreement}
\centering
\setlength{\tabcolsep}{5pt}
\renewcommand{\arraystretch}{1.15}
\begin{tabular}{l c c}
\hline
\textbf{Annotator Pair} & \textbf{Cohen's $\kappa$} & \textbf{Agreement (\%)} \\
\hline
H1 vs.\ H2 & 0.495 & 66.7 \\
H1 vs.\ H3 & 0.471 & 64.3 \\
H2 vs.\ H3 & 0.394 & 59.5 \\
\hline
\textbf{Average} & \textbf{0.453} & \textbf{63.5} \\
\hline
\end{tabular}
\end{table}

\subsubsection{VLM Comparison}

Three frontier VLMs were evaluated on the same 42-pothole sample:
Google Gemini~2.5~Pro, Google Gemini~2.5~Flash, and OpenAI GPT-5. Each
model received identical inputs consisting of the perspective-transformed
scene image, the cropped pothole patch (with dual-image fallback for
distant potholes as described in Section~\ref{sec:stage4}), and the
geometric feature vector from Stage~3. Prompts followed the template of
Appendix~\ref{appendix:depth_prompt} with generation temperature set
to~0.1.

Table~\ref{tab:vlm_comparison} reports each model's agreement with the
consensus-labeled potholes. Because the depth categories are ordinal, we report
quadratic-weighted $\kappa$ alongside unweighted $\kappa$, together with
bootstrap 95\% confidence intervals (5{,}000 scene-level resamples) and
per-class precision and recall (Table~\ref{tab:vlm_perclass}). In this
validation run, where all three models were queried under identical
conditions, Gemini~2.5~Pro attains $\kappa = 0.479$ (weighted
$\kappa = 0.541$, 95\%~CI $[0.28, 0.77]$) with 65.8\% raw agreement, and
Gemini~2.5~Flash and GPT-5 are substantially lower; this comparison is the
basis for the choice of model. The separate full-dataset run that produced the
released ENPAD labels, however, assigns a different class to 8 of the 38
items, and scored against the same panel the released labels give
$\kappa = 0.321$ (weighted $\kappa = 0.406$, 95\%~CI $[0.14, 0.65]$; 55.3\%
agreement). Every downstream figure in this paper uses the released labels,
because those are the labels a reader can reproduce;
Section~\ref{sec:variability} measures the run-to-run variation of the hosted
model.

We do not express these values as a fraction of human agreement. The average
pairwise human $\kappa$ ($\bar{\kappa} = 0.453$) measures agreement among
annotators, not an upper bound on attainable performance. On this panel the
released labels agree with the consensus less closely than the annotators
agree with one another, the confidence intervals are wide, and the panel saw
the models' predictions; Section~\ref{sec:v1} repeats the comparison with a
larger, prediction-free panel.

\begin{table}[!t]
\caption{VLM Agreement with the First Panel's Consensus ($n=38$)}
\label{tab:vlm_comparison}
\centering
\setlength{\tabcolsep}{4pt}
\renewcommand{\arraystretch}{1.15}
\resizebox{\columnwidth}{!}{%
\begin{tabular}{l c c c}
\hline
\textbf{Model} & \textbf{$\kappa$} & \textbf{Weighted $\kappa$ (95\%~CI)}
    & \textbf{Agreement (\%)} \\
\hline
Gemini 2.5 Pro   & \textbf{0.479} & \textbf{0.541} $[0.28, 0.77]$ & \textbf{65.8} \\
Gemini 2.5 Flash & 0.224          & 0.315 $[0.02, 0.58]$          & 50.0 \\
GPT-5            & 0.229          & 0.063 $[-0.02, 0.22]$         & 50.0 \\
\hline
Gemini 2.5 Pro (released labels) & 0.321 & 0.406 $[0.14, 0.65]$ & 55.3 \\
\hline
\multicolumn{4}{l}{\footnotesize Mean pairwise human $\kappa = 0.453$
    (not an upper bound; see text).} \\
\multicolumn{4}{l}{\footnotesize Weighted $\kappa$ is quadratic; CIs are
    bootstrap percentile intervals.} \\
\multicolumn{4}{l}{\footnotesize Rows 1--3: validation run; last row: released
    full-dataset labels.} \\
\end{tabular}}
\end{table}

Inter-model agreement further illustrates the divergence among VLMs.
Gemini~2.5~Flash and Gemini~2.5~Pro showed the highest pairwise
agreement ($\kappa = 0.446$), while Flash versus GPT-5 agreement was
near chance ($\kappa = 0.079$) and Pro versus GPT-5 reached only
$\kappa = 0.270$. These differences likely reflect architectural and
training-data variations across model families, reinforcing the
necessity of empirical model selection rather than assuming equivalent
capability.

Based on its highest agreement in the validation run, Gemini~2.5~Pro was
selected as the severity engine for full-dataset processing of all 3,092
potholes in ENPAD.

\subsection{Warning Generation Ablation Design}
\label{sec:ablation}

\begin{sloppypar}
To quantify the contribution of geometric grounding and risk scoring
to warning quality, we compare two generation variants applied to all
1,119 scenes in ENPAD. Both variants use Google Gemini~2.5~Pro as the
generation backbone and produce the same six-field JSON output schema
(\texttt{warning\_message}, \texttt{maneuver\_suggestion},
\texttt{urgency\_level}, \texttt{reasoning}, \texttt{affected\_lanes},
\texttt{recommended\_speed\_reduction}). The two variants differ mainly in
the information supplied to the model.
\end{sloppypar}

\begin{enumerate}[noitemsep, topsep=3pt]

    \item \textbf{Image-Only Baseline.} The VLM receives only the RGB
    scene image with no geometric features, depth categories, or risk
    scores. The prompt instructs the model to assess road hazards based
    solely on visual observation
    (Appendix~\ref{appendix:image_only_prompt}).

    \item \textbf{Multimodal Hybrid.} The VLM receives the RGB scene
    image together with the complete per-pothole structured assessment
    from Stages~3--4, including depth category and confidence, geometric
    feature values, composite risk score and severity label, a domain
    knowledge rubric defining urgency thresholds, and four worked
    examples demonstrating the mapping from quantified attributes to
    urgency levels (Appendix~\ref{appendix:hybrid_prompt}).

\end{enumerate}

Both variants use the same generative model, but the two runs were not
otherwise matched: the image-only baseline was generated at temperature 0.3
with a looser output schema, and the hybrid with the API's default sampling
settings and a schema that constrains the urgency field. The hybrid prompt also
receives the urgency rubric and worked examples. This comparison therefore does
not by itself isolate the effect of geometric grounding;
Section~\ref{sec:prompt_ablation} repeats it as a six-arm incremental ablation
in which every arm uses one generation configuration. By applying
both variants to every scene rather than a subset, the evaluation
captures performance differences across the full range of scene
complexity, from single shallow defects to multi-pothole high-severity
configurations.

\subsection{Human Evaluation Protocol}
\label{sec:human_eval}

Automated text-quality metrics such as BLEU, ROUGE, and BERTScore
require reference texts and primarily measure surface-level lexical or
semantic overlap, which correlates poorly with the pragmatic qualities
that matter for safety-critical driving advisories, namely clarity of
communication and appropriateness for real-time decision
support~\cite{Reiter2018NLGeval}. We therefore adopt a human evaluation
protocol designed to assess exactly these two dimensions.

\subsubsection{Sample and Annotators}

Fifty scenes were randomly sampled from the 1,119 ENPAD scenes. For
each sampled scene, the warnings produced by both the Image-Only
baseline and the Multimodal Hybrid variant were collected, yielding
100 warning texts in total. Three annotators, rating independently of one
another, participated in the evaluation: two computer science graduate students familiar
with natural language generation research and one professional driver
with over 15~years of driving experience, included to provide a
practitioner perspective on warning utility.

\subsubsection{Double-Blind Rating Procedure}

To eliminate method-identification bias, the two warnings for each
scene were randomly assigned the labels \textbf{A} and \textbf{B},
with the assignment shuffled independently per scene. Annotators were
not informed of which label corresponded to which generation method.
For each scene, every annotator provided three ratings:

\begin{enumerate}[noitemsep, topsep=3pt]

    \item \textbf{Clarity} (1--5 Likert scale): how clear, well
    structured, and easy to understand the warning is.

    \item \textbf{Safety Appropriateness} (1--5 Likert scale): how
    suitable the warning is for safety-critical driving decisions,
    considering urgency calibration, specificity of maneuver advice,
    and avoidance of misleading or vague language.

    \item \textbf{Preference}: which warning the annotator would prefer
    to receive while driving, with three response options (A, B, or
    tie).

\end{enumerate}

This design produces 150 total ratings per Likert dimension (50
scenes $\times$ 3 annotators) and 150 preference judgments. Inter-rater
reliability across the three annotators is measured by Fleiss'
kappa~\cite{Fleiss1971kappa}, which generalizes Cohen's $\kappa$ to
three or more raters. For $n$ items rated by $m$ raters into $k$
categories, Fleiss' kappa is defined as:
\begin{equation}
    \kappa_F = \frac{\bar{P} - \bar{P}_e}{1 - \bar{P}_e},
    \label{eq:fleiss_kappa}
\end{equation}
where $\bar{P}$ is the mean proportion of pairwise agreement across
all items and $\bar{P}_e$ is the expected agreement under random
assignment, computed from the overall category
proportions~\cite{Fleiss1971kappa}. This metric assesses the
consistency of human judgment on both Likert scales and the preference
dimension.

\subsubsection{Statistical Analysis}
\label{sec:stats}

Because every scene is rated by the same three annotators under both methods,
the 150 ratings are not independent observations. We therefore analyze rating
differences at the scene level: the hybrid-minus-image-only difference is
averaged over annotators within each scene ($n = 50$) and tested with the
Wilcoxon signed-rank test, and 95\% confidence intervals are obtained from a
bootstrap that resamples whole scenes (5{,}000 resamples). We repeat the paired
test separately for each annotator. Preferences are analyzed by scene: each
scene is assigned to the method that received more of its three votes, and an
exact binomial test is applied to the scenes that are not tied. The urgency
labels of the two methods are paired on the same 1,119 scenes, so their
distributions are compared with the Stuart--Maxwell test of marginal
homogeneity. Agreement statistics use Cohen's or Fleiss' $\kappa$, with
quadratic weights where categories are ordinal, and bootstrap percentile
intervals.

\section{Results}
\label{sec:results}

This section presents the empirical evaluation of the proposed hybrid
warning generation pipeline against the image-only baseline across all
1,119 scenes in the dataset. The analysis proceeds in four parts. First,
we examine the distributional shift in warning urgency levels produced by
each method. Second, we quantify the degree of agreement and systematic
divergence between the two methods. Third, we report the human evaluation
study results, including inter-rater reliability, perceptual quality
ratings, and overall method preference. Fourth, we present qualitative
side-by-side examples that illustrate, at the scene level, how the
geometric grounding reshapes the language and specificity of the
generated warnings.

\subsection{Warning Urgency Distribution Analysis}
\label{sec:urgency_distribution}

Table~\ref{tab:urgency_distribution} summarizes the four-level urgency
distributions produced by the image-only and hybrid pipelines across all
1,119 scenes. The image-only method concentrates the majority of its
outputs in the high-urgency tier, with 65.7\% of scenes classified as
high or critical (658 high, 77 critical). In contrast, the hybrid method
distributes its outputs more broadly, assigning only 41.6\% of scenes to
the combined high and critical tiers (304 high, 161 critical). This
24.1~percentage-point reduction in high-or-critical assignments shows that
the hybrid method assigns the top two tiers far less often than the image-only
method.

\begin{table}[!t]
\caption{Warning Urgency Distribution Comparison ($n=1{,}119$ scenes)}
\label{tab:urgency_distribution}
\centering
\setlength{\tabcolsep}{5pt}
\renewcommand{\arraystretch}{1.2}
\begin{tabular}{l r@{\,}l r@{\,}l}
\toprule
\textbf{Urgency Level} & \multicolumn{2}{c}{\textbf{Image-Only}} & \multicolumn{2}{c}{\textbf{Hybrid}} \\
\midrule
Low      & 78  & (7.0\%)  & 58  & (5.2\%)  \\
Medium   & 306 & (27.3\%) & 596 & (53.3\%) \\
High     & 658 & (58.8\%) & 304 & (27.2\%) \\
Critical & 77  & (6.9\%)  & 161 & (14.4\%) \\
\midrule
\multicolumn{5}{l}{Stuart--Maxwell $\chi^{2} = 260.1$, df $= 3$, $p = 4.2\times10^{-56}$} \\
\bottomrule
\end{tabular}
\end{table}

The redistribution is not uniform across all tiers. The most pronounced
shift occurs between the high and medium categories, where 299 scenes
that the image-only pipeline assigned to high urgency are downgraded to
medium by the hybrid method. This 26.7\% reduction in high-urgency
assignments is a change in the urgency distribution rather than a measured
false-alarm rate: no independently labeled hazard ground truth exists for
these scenes, so the two methods can be compared with each other but not
against an external notion of correctness. The shift reflects the geometric
features' capacity to
contextualize visual severity cues with positional information; potholes
that appear visually severe but lie outside the vehicle's wheel path or
at a greater distance are assigned lower urgency when spatial context is
available.

\begin{figure}[!t]
    \centering
    \includegraphics[width=\columnwidth]{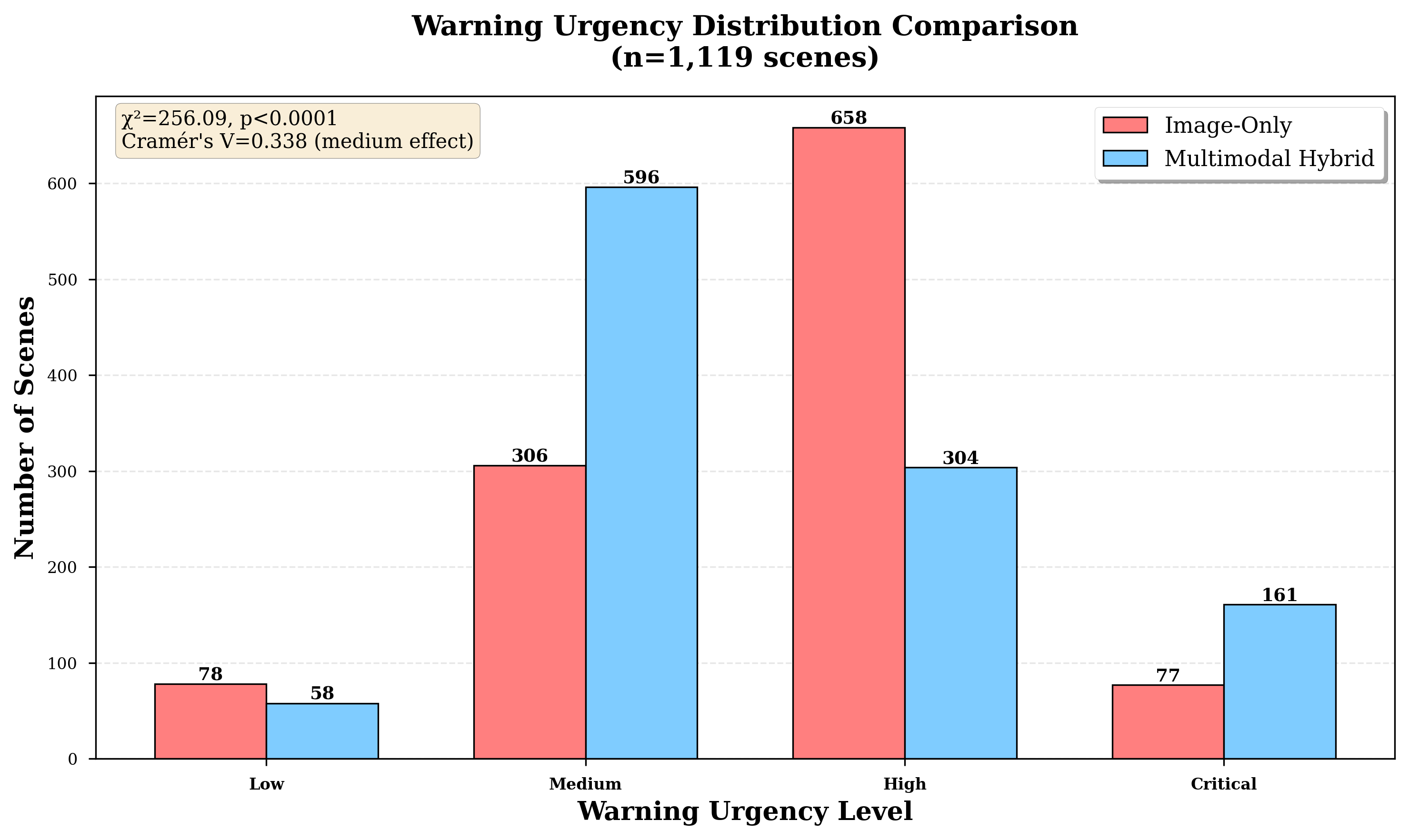}
    \caption{Warning urgency distribution across all 1,119 scenes for the
    image-only and hybrid methods. The hybrid pipeline shifts a substantial
    portion of high-urgency assignments to medium while increasing
    critical-tier assignments.}
    \label{fig:urgency_distribution}
\end{figure}

At the upper extreme, the hybrid pipeline assigns the critical tier to
109\% more scenes than the image-only baseline (161 vs.\ 77 scenes). We
describe this as an increase in critical-tier assignments rather than as
improved detection of genuine hazards, since no independent hazard labels
are available. Inspection of the affected scenes indicates that the
upgraded cases are predominantly large, deep potholes lying directly in the
vehicle's wheel path at close range, which the image-only variant does not
separate from the broader pool of high-urgency warnings.
Fig.~\ref{fig:urgency_distribution} visualizes the distributional
differences between the two methods.

Because both methods are applied to the same 1,119 scenes, the two sets of
urgency labels are paired and a test of independence is inappropriate. We
therefore use the Stuart--Maxwell test of marginal homogeneity, which
rejects the hypothesis that the two methods induce the same urgency
distribution ($\chi^{2} = 260.1$, $\mathrm{df} = 3$, $p = 4.2\times
10^{-56}$). The redistribution is large: 55.4\% of scenes receive a
different urgency level from the two methods, and the changes follow the
consistent pattern described below rather than scattering randomly.

\subsection{Inter-Method Agreement}
\label{sec:inter_method}

To characterize the nature of the divergence between the two warning
pipelines, we compute pairwise agreement statistics across the full
dataset. Table~\ref{tab:method_agreement} reports the results. Cohen's
$\kappa = 0.186$ indicates only slight agreement between the two
methods according to the standard interpretation
scale~\cite{Landis1977kappa}, while exact agreement, defined as
both methods assigning the identical urgency level, occurs in only
44.6\% of scenes. When agreement is relaxed to include adjacent
categories (e.g., high vs.\ critical or medium vs.\ high), the rate
rises to 95.4\%.

\begin{table}[!t]
\caption{Agreement Between Warning Methods ($n=1{,}119$ scenes)}
\label{tab:method_agreement}
\centering
\setlength{\tabcolsep}{5pt}
\renewcommand{\arraystretch}{1.2}
\begin{tabular}{l l}
\toprule
\textbf{Metric} & \textbf{Value} \\
\midrule
Cohen's Kappa ($\kappa$)   & 0.186 (Slight) \\
Exact Agreement            & 44.6\% \\
Adjacent Agreement         & 95.4\% \\
Stuart--Maxwell $\chi^{2}$ & 260.1 (df $= 3$), $p < 10^{-15}$ \\
\bottomrule
\end{tabular}
\end{table}

The low $\kappa$ value does not indicate random or unreliable
disagreement. Rather, it reflects a systematic distributional shift:
the hybrid method consistently reassigns scenes to different urgency
tiers based on the additional geometric context. Analysis of the
confusion structure reveals two dominant transition patterns. First,
299 scenes are downgraded from high to medium, accounting for the
majority of the redistribution. These cases typically involve potholes
that appear visually severe in the image but occupy positions outside
the direct wheel path or at distances where immediate action is
unnecessary. Second, 110 scenes are upgraded from high to critical,
corresponding to potholes whose geometric features, particularly large
area, low circularity, and in-path lateral position, indicate hazards
that demand immediate driver attention. The combination of high adjacent
agreement (95.4\%) with low exact agreement (44.6\%) indicates that the two
methods mostly differ by one urgency level, in a consistent direction. Without
independent hazard labels we cannot say which method's level is correct for a
given scene.

\subsection{Human Evaluation Results}
\label{sec:human_eval_results}

\subsubsection{Inter-Rater Reliability}

Before analyzing the comparative ratings, we assess the consistency of
the three annotators using Fleiss' kappa~\cite{Fleiss1971kappa}.
Table~\ref{tab:irr} reports the reliability coefficients for each
method-criterion combination, computed per method after the A/B labels are
mapped back to the methods. Kappa values range from 0.458 to 0.603, with an
average of $\kappa = 0.553$, which corresponds to moderate
agreement~\cite{Landis1977kappa}: the annotators applied the criteria
consistently enough to support between-method comparisons.

\begin{table}[!t]
\caption{Inter-Rater Reliability (Fleiss' Kappa)}
\label{tab:irr}
\centering
\setlength{\tabcolsep}{5pt}
\renewcommand{\arraystretch}{1.2}
\begin{tabular}{l c l}
\toprule
\textbf{Method / Criterion} & \textbf{$\kappa$} & \textbf{Interpretation} \\
\midrule
Image-only -- Clarity & 0.590 & Moderate \\
Image-only -- Safety  & 0.561 & Moderate \\
Hybrid -- Clarity     & 0.458 & Moderate \\
Hybrid -- Safety      & 0.603 & Moderate \\
\midrule
Average               & 0.553 & Moderate \\
\bottomrule
\end{tabular}
\end{table}

Reliability is similar across methods and criteria; it is lowest for the
clarity of the hybrid warnings ($\kappa = 0.458$), whose additional technical
detail may be judged differently by evaluators with different backgrounds.

\subsubsection{Perceptual Quality Ratings}

Table~\ref{tab:human_eval} presents the mean ratings and the scene-level
paired comparisons described in Section~\ref{sec:stats}. The hybrid pipeline
achieves a mean clarity rating of $4.04 \pm 0.87$ compared to
$3.21 \pm 1.04$ for the image-only method, a difference of $+0.83$ points
(95\%~CI $[0.54, 1.11]$, scene-level Wilcoxon $p = 6.6\times10^{-6}$). For
safety relevance, the hybrid method scores $3.98 \pm 1.01$ versus
$3.25 \pm 1.12$, a difference of $+0.73$ points ($[0.36, 1.11]$,
$p = 7.2\times10^{-4}$). All three annotators show the same direction and a
similar size of effect (clarity $+0.86$, $+0.80$, $+0.82$; safety $+0.66$,
$+0.80$, $+0.72$; per-annotator Wilcoxon $p \leq 0.0023$). We report the
differences with their confidence intervals rather than against a fixed
threshold, since no established minimum important difference exists for this
rating task. One safety rating was entered as 6 on the 1--5 scale; we keep it
as recorded, and capping it at 5 changes the safety difference from $+0.73$ to
$+0.72$.

\begin{table}[!t]
\caption{Human Evaluation Results ($n=50$ scenes, 3 annotators, 150
total ratings). CI: scene-clustered bootstrap; $p$: scene-level Wilcoxon
signed-rank test.}
\label{tab:human_eval}
\centering
\setlength{\tabcolsep}{4pt}
\renewcommand{\arraystretch}{1.2}
\setlength{\tabcolsep}{2.5pt}
\begin{tabular}{l c c c c}
\toprule
\textbf{Criterion} & \textbf{Image-Only} & \textbf{Hybrid} & \textbf{Diff. [95\% CI]} & \textbf{$p$} \\
\midrule
Clarity   & $3.21 \pm 1.04$ & $4.04 \pm 0.87$ & $+0.83$ $[0.54, 1.11]$ & $6.6\!\times\!10^{-6}$ \\
Safety    & $3.25 \pm 1.12$ & $3.98 \pm 1.01$ & $+0.73$ $[0.36, 1.11]$ & $7.2\!\times\!10^{-4}$ \\
\bottomrule
\end{tabular}
\end{table}

The clarity improvement suggests that the geometric context enables the
VLM to generate warnings with more specific spatial references, such as
describing a pothole's position relative to the wheel path or its
proximity to the vehicle, rather than relying solely on generic visual
severity descriptors. The safety relevance improvement, while slightly
smaller in magnitude, indicates that human evaluators perceived the
hybrid warnings as more closely aligned with the actual driving hazard
present in each scene.

\subsubsection{Overall Method Preference}

Table~\ref{tab:preference} reports the forced-choice preference
distribution across all 150 evaluator-scene pairs. The hybrid method
was preferred in 92 of 150 comparisons (61.3\%), compared to 26
preferences for the image-only method (17.3\%) and 32 tie judgments
(21.3\%), a preference ratio of 3.5:1 in favor of the hybrid approach
(95\%~CI $[2.0, 7.5]$, scene-clustered bootstrap). Analyzed by scene, the
hybrid warning received more votes in 34 scenes and the image-only warning in
9, with 7 scenes tied (exact binomial test on the 43 untied scenes,
$p = 1.7\times10^{-4}$). Each annotator individually preferred the hybrid
warnings (29:6, 33:10 and 30:10, excluding ties).

\begin{table}[!t]
\caption{Overall Method Preference ($n=150$ votes)}
\label{tab:preference}
\centering
\setlength{\tabcolsep}{5pt}
\renewcommand{\arraystretch}{1.2}
\begin{tabular}{l r@{\,}l}
\toprule
\textbf{Preference} & \multicolumn{2}{c}{\textbf{Votes}} \\
\midrule
Hybrid (Method B)     & 92  & (61.3\%) \\
Image-Only (Method A) & 26  & (17.3\%) \\
Tie                   & 32  & (21.3\%) \\
\midrule
\multicolumn{3}{l}{Ratio $3.5\times$ $[2.0, 7.5]$; scene-level binomial $p = 1.7\times10^{-4}$} \\
\bottomrule
\end{tabular}
\end{table}

\begin{figure}[!t]
    \centering
    \includegraphics[width=\columnwidth]{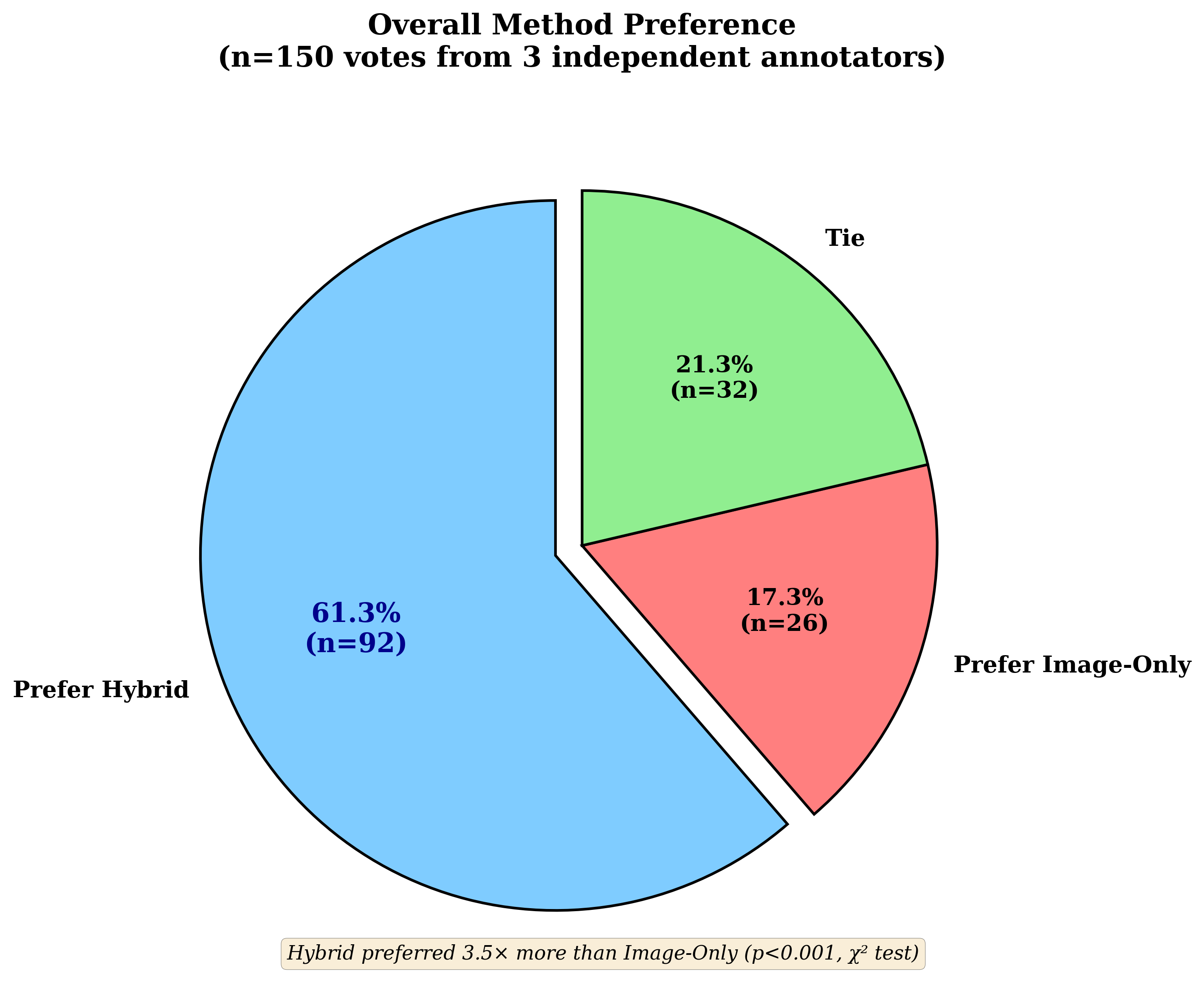}
    \caption{Distribution of annotator preferences across 150
    evaluator-scene pairs. The hybrid method was preferred in 61.3\% of
    comparisons, with a 3.5:1 preference ratio over the image-only
    baseline.}
    \label{fig:preference_distribution}
\end{figure}

The 3.5:1 preference ratio is particularly notable given that the
evaluation was conducted in a blinded setting where annotators did not
know which method produced each warning, which removes knowledge of the
method as a source of bias. The 21.3\% tie rate indicates that in roughly
one-fifth of comparisons both methods produce warnings of comparable quality;
geometric grounding may help most where spatial context is ambiguous from
visual information alone.
Fig.~\ref{fig:preference_distribution} visualizes the preference
distribution across all annotator-scene pairs.

\subsection{Qualitative Examples}
\label{sec:qualitative_examples}

\begin{figure}[!t]
    \centering
    \includegraphics[width=\columnwidth]{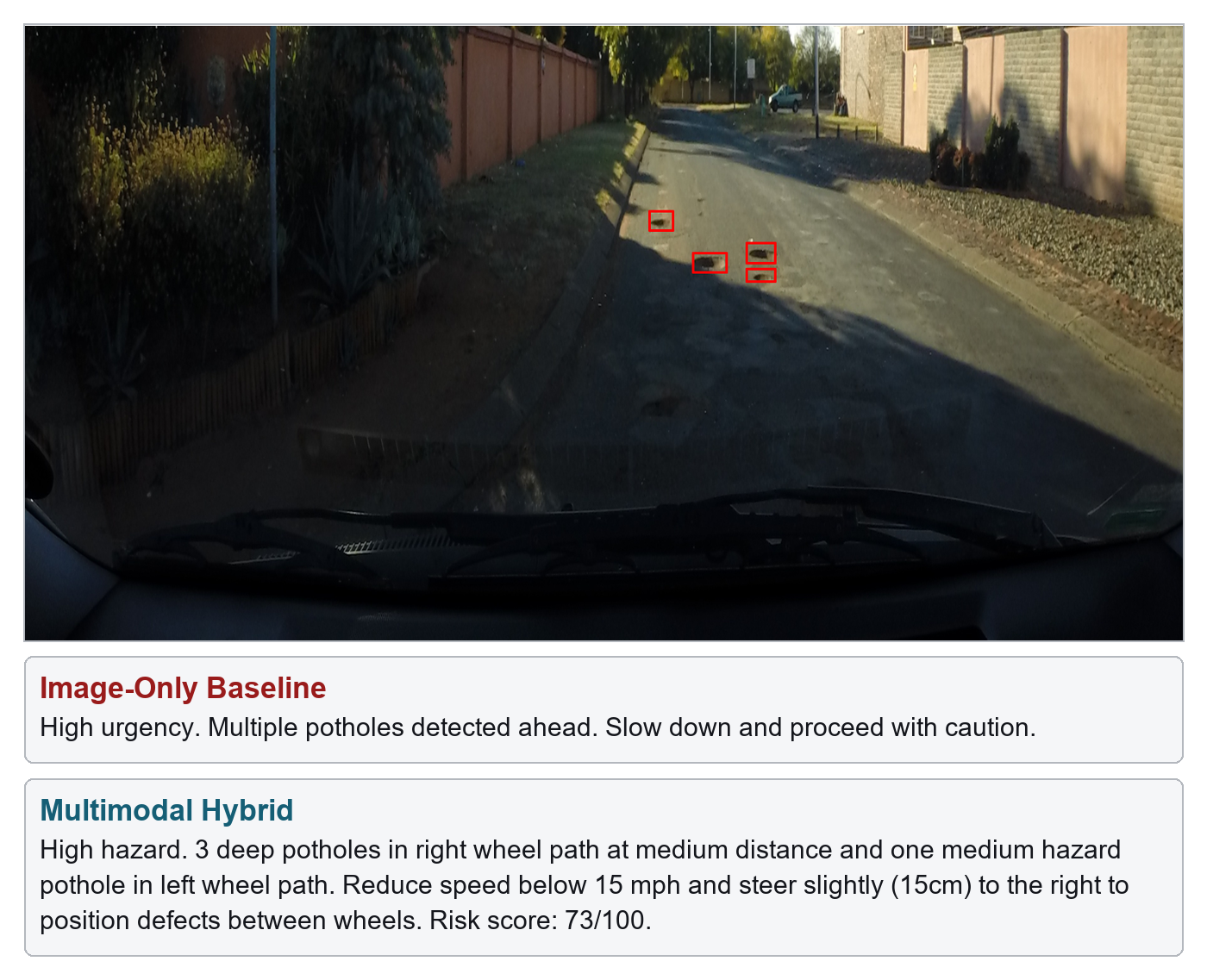}
    \caption{Qualitative warning-generation example for a high-hazard
    scene. Top: dashcam image with annotated pothole bounding boxes.
    Middle: warning produced by the image-only baseline. Bottom:
    warning produced by the multimodal hybrid pipeline, which combines
    the same visual input with the deterministic geometric feature
    vector. The hybrid output localizes hazards to wheel paths,
    recommends a specific steering offset, and reports a composite
    risk score.}
    \label{fig:warning_example1}
\end{figure}

\begin{figure}[!t]
    \centering
    \includegraphics[width=\columnwidth]{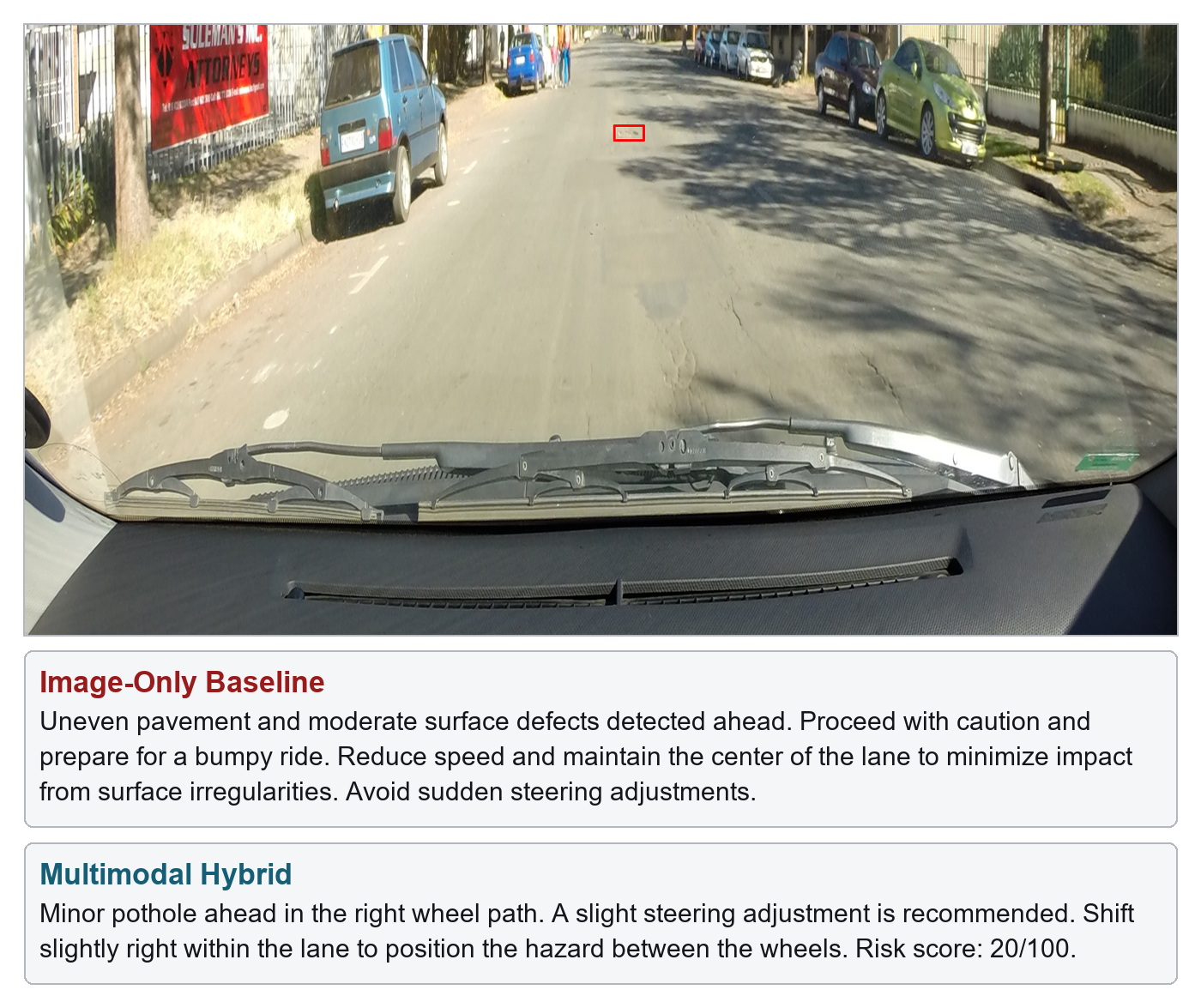}
    \caption{Qualitative warning-generation example for a minor-hazard
    scene with a single small defect. Top: dashcam image with annotated
    pothole bounding box. Middle: warning produced by the image-only
    baseline, which describes the scene in generic terms. Bottom:
    warning produced by the multimodal hybrid pipeline, which
    identifies the defect's wheel-path location, suggests a slight
    lane-internal steering adjustment, and assigns a low composite
    risk score.}
    \label{fig:warning_example2}
\end{figure}

To complement the aggregate statistics with concrete scene-level
evidence, Fig.~\ref{fig:warning_example1} and
Fig.~\ref{fig:warning_example2} present two representative dashcam
scenes alongside the warnings produced by the image-only baseline and
the multimodal hybrid pipeline. In both cases, the two methods receive
identical visual input; only the hybrid pipeline additionally consumes
the deterministic geometric feature vector extracted in Stage~3.

Fig.~\ref{fig:warning_example1} corresponds to a high-hazard scene in
which several potholes lie ahead of the vehicle. The image-only
baseline produces a generic alert that flags the presence of multiple
potholes and recommends slowing down, but provides no spatial detail
and no quantitative severity estimate. The hybrid pipeline, in
contrast, localizes the defects to specific wheel paths, distinguishes
the dominant cluster on the right from a secondary defect on the
left, suggests a concrete steering offset, and reports a composite
risk score. The added specificity directly reflects the geometric
features made available to the language model.

Fig.~\ref{fig:warning_example2} illustrates the opposite end of the
hazard spectrum. The image-only baseline interprets the scene as
broadly uneven pavement and issues a cautious, somewhat generic
recommendation. The hybrid pipeline reports a single
minor pothole in the right wheel path, recommends a small steering
shift to bracket the defect between the wheels, and reports a low
risk score. This example is consistent with the broader pattern
observed in Section~\ref{sec:urgency_distribution}, in which the
hybrid method downgrades many visually salient but low-impact scenes
from high to medium urgency while still issuing an actionable, lane-
aware maneuver suggestion.

Taken together, the quantitative distribution analysis, inter-method
agreement statistics, and human evaluation results converge on a
consistent finding. Integrating geometric features into warning generation
produces a systematic shift in urgency levels and a human-perceptible
improvement in the rated clarity and safety relevance of the warnings. Whether
the shifted urgency levels are more correct cannot be determined without
independent hazard labels.

\section{Validation Against Independent References and Robustness Analyses}
\label{sec:new_analyses}

The evaluations of Sections~\ref{sec:experiments} and~\ref{sec:results} compare
the pipeline either with a human panel that saw the models' predictions or with
its own image-only variant. This section adds references that are independent
of the pipeline, and tests how robust its outputs are.

\subsection{Validation Against Physically Measured Depth}
\label{sec:metric_validation}

The source dataset carries no metric depth annotation. To test whether the
severity classes correspond to physical depth, we use
PothRGBD~\cite{pothrgbd, yurdakul2025}, an independent public RGB-D pothole
dataset with co-registered depth in millimeters and human
polygon annotations, which takes no part in the main study. For each annotated
pothole we fit a robust plane to the surrounding intact road surface, compute
the tilt-corrected deviation of the interior from that plane, and take the 90th
percentile as the physical depth. This yields metric ground truth for 910
potholes (median 23.7\,mm, interdecile range 7.9--43.4\,mm; median plane-fit
residual 5.1\,mm, i.e.\ the surface model is markedly tighter than the depths
being measured). Applying the class thresholds gives 497 medium, 370 low, and 43
high potholes.

We then ran the Stage-4 engine, with the prompt of
Appendix~\ref{appendix:depth_prompt}, on RGB crops from this dataset
($n = 151$, sampled to balance the three metric classes) and compared its output
with the sensor-derived class (Table~\ref{tab:metric_validation}). Agreement
with physical depth is weak: $\kappa = 0.098$ (quadratic-weighted
$\kappa = 0.178$, 95\% CI $[0.05, 0.30]$), with 42.4\% exact agreement,
although 88.7\% of predictions fall within one class of the measured class.

\begin{table}[!t]
\caption{Stage-4 severity classes against sensor-measured depth on PothRGBD
($n = 151$). Rows: measured class; columns: predicted class. $\kappa = 0.098$;
weighted $\kappa = 0.178$ $[0.05, 0.30]$; Spearman $\rho = 0.214$, $p = 0.009$.}
\label{tab:metric_validation}
\centering
\setlength{\tabcolsep}{5pt}
\resizebox{\columnwidth}{!}{%
\begin{tabular}{l c c c c}
\hline
\textbf{Measured} & \textbf{Pred.\ low} & \textbf{Pred.\ medium} & \textbf{Pred.\ high} & \textbf{Recall} \\
\hline
Low ($<$20\,mm)     & 36 & 18 & 1 & 0.655 \\
Medium (20--50\,mm) & 29 & 25 & 1 & 0.455 \\
High ($>$50\,mm)    & 16 & 22 & 3 & 0.073 \\
\hline
Mean measured depth & 28.4\,mm & 36.1\,mm & 44.1\,mm & \\
\hline
\end{tabular}}
\end{table}

Two features of this result matter. First, the ordering is preserved: mean
measured depth rises monotonically across the predicted classes, and the rank
correlation between predicted class and measured depth is significant
($\rho = 0.214$, $p = 0.009$). Second, the model is strongly conservative at
the deep end: it assigned the high class to only 5 of 151 potholes, recovering
7\% of those measuring more than 50\,mm.

The Stage-4 output therefore carries ordinal information about physical depth
but is not a calibrated measurement of it, and it systematically under-calls
the deepest defects. This is the evidential basis for the terminology used
throughout this paper: the classes are \emph{perceived visual severity
classes} named after maintenance thresholds, not depth measurements. Two
caveats bound the conclusion. PothRGBD is captured at close range rather than
from a dashcam viewpoint, so the visual cues differ from the operating regime
of the main study; and the distance and lateral-position fields of the prompt,
which carry information in the dashcam setting, are uninformative here. The
finding should therefore be read as evidence about the severity criterion
itself rather than about dashcam performance.

\subsection{Incremental Prompt Ablation}
\label{sec:prompt_ablation}

The comparison of Section~\ref{sec:results} contrasts an image-only prompt with
the full hybrid prompt, which also supplies the decision rubric and worked
examples, and the two published runs also differed in sampling settings and
output schema (Section~\ref{sec:ablation}). To separate the contribution of
each information component under controlled conditions, we ran six arms with
Gemini~2.5~Pro on the same 300 randomly sampled scenes, using one generation
configuration, that of the published hybrid run, for every arm. Arm~1 is the
published image-only prompt and arm~6 the published hybrid prompt, both
verbatim; arms~2--5 are obtained by deleting components from the hybrid prompt
(Table~\ref{tab:prompt_ablation}). All 1,800 calls returned valid outputs.

\begin{table}[!t]
\caption{Incremental prompt ablation with Gemini~2.5~Pro ($n = 300$ scenes, one
generation configuration for all arms). Rubric consistency is agreement between
the generated urgency and the deterministic risk-score rubric.}
\label{tab:prompt_ablation}
\centering
\setlength{\tabcolsep}{4pt}
\resizebox{\columnwidth}{!}{%
\begin{tabular}{l c c c}
\hline
\textbf{Prompt arm} & \textbf{High+Critical (\%)} & \textbf{Rubric cons.\ (\%)} & \textbf{Agree w/ full (\%)} \\
\hline
1. Image only (published) & 56.7 & 44.7 & 45.3 \\
2. \; + rubric & 42.0 & 46.7 & 47.0 \\
3. \; + rubric + size, position & 78.7 & 31.7 & 32.3 \\
4. \; + rubric + severity, position & 33.7 & 76.3 & 75.7 \\
5. \; + all data, no examples & 38.7 & 99.0 & 98.3 \\
6. Full hybrid (published) & 39.0 & 99.3 & 100.0 \\
\hline
\end{tabular}}
\end{table}

Supplying the rubric alone changes rubric consistency by only 2.0 percentage
points relative to the image-only prompt (44.7\% to 46.7\%), although it lowers
high-or-critical assignments from 56.7\% to 42.0\%; adding the per-pothole
assessment on top of the rubric raises consistency by a further 52.3 points, to
99.0\%. Removing the four worked examples changes consistency by only 0.3
points. The hybrid method's rule-adherence is therefore not produced by being
given the rules: it comes from the per-pothole measurements, with the severity
class as the strongest single component (76.3\% with severity and position
alone), while size and position without severity lead to heavy over-alerting
(78.7\% high or critical). The full hybrid arm closely reproduces the published
hybrid run (39.0\% high or critical here, against 41.6\% over all 1,119 scenes
in Table~\ref{tab:urgency_distribution}). Two qualifications apply. Arms that
receive the composite risk score can satisfy the rubric more easily by
construction, so absolute values are not directly comparable across arms with
and without that field. And all arms are scored against pipeline-derived
references, so the ablation identifies which information changes the output,
not which output is correct; that requires independently created hazard and
maneuver labels.

\subsection{Risk-Score Sensitivity and Scoring Reconciliation}
\label{sec:risk_sensitivity}

We perturbed each weight by $\pm 0.1$ over the simplex and each category
boundary by $\pm 5$ points. Category assignments change for 9.5--43.8\% of
potholes under weight perturbation and 19.6--25.5\% under boundary shifts.
Two properties qualify these figures. First, every change is to an adjacent
band; no pothole is reclassified non-adjacently under any perturbation. Second,
the induced ranking is essentially invariant (Spearman $\rho \geq 0.954$
against the reference score). The numeric boundaries are therefore best
understood as operating-point conventions on a continuous score: which side of
a boundary a borderline pothole falls on is sensitive to the weights, whereas
the prioritization of potholes, which is what maintenance triage uses, is not.
These results concern stability only; whether the score tracks actual damage
risk cannot be tested without outcome data.

The scorer of Table~\ref{tab:risk_weights} reproduces the released risk
categories for 99.8\% of potholes. An alternative formulation with
percentile-binned area scores and a distance-dependent position score agrees
with it exactly for 66.1\% of potholes and within one band for 99.9\%
($\rho = 0.911$). Percentile bins computed on a 40\% calibration split change
1.2\% of categories on the held-out 60\%; the scorer itself uses fixed
constants and no data-derived thresholds. Finally, the numeric bands and the
trigger conditions of Table~\ref{tab:urgency_rubric} agree for 92.6\% of
potholes; the numeric band is primary and the trigger conditions act as
escalation-only rules.

\subsection{Measured Segmentation Quality}
\label{sec:seg_quality}

SAM's self-reported IoU is a confidence estimate, not a measurement against a
reference. We therefore evaluated the identical box-prompted protocol against
126 human-drawn polygon masks from the independent RGB-D dataset. Measured IoU
is 0.780 on average (median 0.809; 96.0\% of masks above 0.5, 67.5\% above
0.75), whereas the model's self-reported IoU averages 0.953, an overestimate of
0.173, with a correlation of only $r = 0.569$ between the two. Segmentation
quality is good, but the predicted score should not be read as segmentation
accuracy.

We also quantified the behavior of monocular foundation depth models
illustrated in Fig.~\ref{fig:depth_ablation}. Using the same RGB-D dataset, we
compared the interior-versus-road depth step measured by the sensor with the
step predicted by the Small variant of Depth Anything~v2 over 95 potholes. The
sensor records a median step of 17.0\,mm, while the model's predicted step lies
within one standard deviation of the local surface noise for 85.3\% of
potholes, and its correlation with the physical step is $r = 0.005$.

\subsection{Repeated-Run Variability}
\label{sec:variability}

A low generation temperature does not make a hosted model deterministic.
Repeating severity classification three times over 80 potholes at the
temperature of 0.1 yields identical classes across all runs for 91.3\% of
potholes (Fleiss' $\kappa = 0.815$ across runs), and a present-day run agrees
with the released labels for 88.8\% of them. This bounds the reproducibility
of any single run and is why the complete generated outputs are kept alongside
the code.

\subsection{Expanded Prediction-Free Human Evaluation}
\label{sec:v1}

The first depth panel (Section~\ref{sec:human_validation}) has three
weaknesses: 38 usable items, an annotation sheet that displayed the models'
predictions, and a mean pairwise agreement of $\kappa = 0.453$. We therefore
collected a second, larger panel under stricter conditions.

\textbf{Protocol.} The panel covers 102 potholes: the 42 of the first panel,
plus 60 new scene-disjoint samples drawn 20/20/20 across the pipeline's
severity labels for class balance (the labels themselves were never shown).
Three volunteer annotators, working independently, saw only the defect crop and
the full scene. \textbf{No model predictions, geometric features, or
confidences appeared in the interface.} Each annotator recorded a severity
class, a difficulty flag, and free-text notes. Before analysis we checked the
returned files for signs of automated completion (file metadata, editing
history, and agreement patterns); they showed none.

\textbf{Reliability.} Fleiss $\kappa = 0.604$; mean pairwise $\kappa = 0.605$
(range 0.519--0.759); 63 of 102 items unanimous; a single three-way split,
which we exclude rather than tie-break. Unanimity is 76.5\% on items all three
annotators marked \emph{easy} and 52.0\% on items all three marked
\emph{hard}: difficulty predicts disagreement, as it should.

\textbf{Result.} Table~\ref{tab:v1} reports the pipeline's severity agreement
against this reference.

\begin{table}[!t]
\caption{Pipeline severity agreement (released labels) against the two human
panels. The second row is the like-for-like comparison with the first panel.}
\label{tab:v1}
\centering
\small
\setlength{\tabcolsep}{3pt}
\begin{tabular}{lccc}
\toprule
Reference & $\kappa$ & weighted $\kappa$ & human $\bar{\kappa}$ \\
\midrule
First panel, 38 items & 0.321 & 0.406 & 0.453 \\
Second panel, same 42 items & 0.218 & 0.342 & 0.759 \\
Second panel, 59 new items & 0.090 & 0.328 & 0.485 \\
\textbf{Second panel, all 101 items} & \textbf{0.139} & \textbf{0.320} & \textbf{0.605} \\
\bottomrule
\multicolumn{4}{l}{\footnotesize All 101 items: $\kappa$ 95\% CI $[0.00, 0.28]$; human $\bar{\kappa}$ over all 102 items.} \\
\end{tabular}
\end{table}

\textbf{Why agreement fell.} The decline from 0.321 to 0.139 has two sources.
On the same 42 potholes, the engine's agreement falls from 0.321 against the
first panel to 0.218 against the second, while agreement among annotators rises
from 0.453 to 0.759: the reference became more reliable and moved away from the
engine. The 59 new, class-balanced items are harder for both people
($\bar{\kappa} = 0.485$) and the engine ($\kappa = 0.090$), so part of the
overall decline also reflects the harder sample. The two panels' consensus
labels agree with one another at only $\kappa = 0.378$ on the 38 items both
define.

\textbf{Interpretation.} Three references now bear on the same question:
physically measured depth on an independent RGB-D dataset ($\kappa = 0.098$),
the expanded prediction-free human panel ($\kappa = 0.139$), and the first,
prediction-displaying panel ($\kappa = 0.321$ for the released labels, 0.479
for the validation run). The first two agree; the third is the outlier and is
the one collected with model predictions visible. We conclude that monocular
categorical severity classification, as implemented here, agrees only weakly
with both physical measurement and careful human judgment.

This does not affect the paper's other findings, which never depended on the
human depth reference: the 26.7\% urgency-tier shift and 109\% critical-tier
increase are measured against the pipeline's own image-only baseline, and the
3.5:1 blinded preference for grounded warnings is a separate study with its own
annotators.

\section{Discussion}
\label{sec:discussion}

\subsection{Offline Assessment vs.\ Real-Time Deployment}
\label{sec:offline_vs_realtime}

The proposed framework is designed and evaluated as an \emph{offline}
road infrastructure assessment tool applied to recorded dashcam
footage, not as a real-time perception system. End-to-end processing
of approximately 18--23\,s per scene reflects this scope: the dominant
contributors are SAM~ViT-H inference, which requires several seconds
per frame on consumer-grade hardware~\cite{Kirillov2023}, and the Gemini API
calls (one severity call per pothole and one warning call per scene), each
incurring network round-trip overhead. These components were selected to
maximize assessment quality, and for the targeted offline use case, where
dashcam footage is batch-processed after collection, per-frame latency is
secondary to assessment quality.

The modular architecture nonetheless provides a clear pathway toward
real-time adaptation: each stage can be optimized or replaced
independently. Stage~1 (perspective transformation) and Stage~3
(geometric feature extraction) are already deterministic and execute
in under 10\,ms combined. Stage~2 (SAM) and the API-bound components
of Stage~4--5 are the natural targets for engineering optimization;
lightweight segmentation backbones and on-device distillation of the
depth and warning components are discussed in
Section~\ref{sec:future_work}. The contribution of this work is
therefore methodological: it shows how structured geometric context changes
VLM-generated warnings, and it measures the limits of VLM severity
classification against independent references.

\subsection{Over-Conservatism in Vision-Only Systems}
\label{sec:over_conservatism}

The ablation reveals a systematic pattern of over-conservative risk
classification when VLMs operate on visual information alone. The
image-only baseline classified 65.7\% of scenes as high or critical
risk, compared to 41.6\% under the geometrically grounded pipeline.
Shadows, water reflections, lighting variations, and surface
discoloration can all mimic the appearance of deep potholes, and
without explicit geometric constraints the VLM tends toward worst-case
interpretations under visual ambiguity. A likely practical consequence is
alarm fatigue: when most warnings sit at the highest urgency, the
system loses its discriminative value and drivers desensitize to
legitimate alerts~\cite{Lim2023HMI, Louw2020}, a pattern also
observed in production ADAS phantom-braking
events~\cite{TESMAG2025, Timm2025}.

Geometric grounding addresses this without uniformly attenuating risk.
Critical-tier assignments rose from 77 to 161 scenes while the bulk of
high-tier assignments shifted to medium, indicating a \emph{selective
redistribution}: scenes dominated by visually salient but small or
out-of-path defects are downgraded, while large, deep, in-path potholes are
elevated. This pattern is the behavior one would want from a warning system,
but without independent hazard labels we cannot confirm that the individual
reassignments are correct. We describe it as a redistribution rather than as a change in sensitivity or
specificity, since those quantities require independent hazard labels that do
not exist for this dataset.

\subsection{The Role of Structured Information in VLM Reasoning}
\label{sec:structured_info}

The results indicate that VLM-generated warnings change substantially when
explicitly structured geometric context is supplied, and the incremental
ablation (Section~\ref{sec:prompt_ablation}) attributes rule-adherence to the per-pothole measurements rather than to the rubric text
or worked examples.
This benefit concerns the warnings; it does not extend to the severity classes
themselves, which agree only weakly with independent references
(Section~\ref{sec:depth_limitations}). Rather than treating the VLM as a standalone
visual oracle, the framework positions it as one component in a
hybrid architecture in which classical computer vision supplies
reproducible measurements and the VLM contributes flexible
scene interpretation and natural-language generation.

The geometric features extracted in Stage~3 act as ``anchors'' that
constrain the VLM's interpretation space: with explicit information
about mask area, circularity, aspect ratio, and spatial position, the
model can no longer speculate freely from ambiguous visual cues alone.
This anchoring effect is consistent with prior findings that spatial
grounding significantly improves VLM performance on geometric reasoning
tasks~\cite{SpatialRGPT2024}. By fusing the two modalities through a
structured JSON interface rather than end-to-end learned
representations, the framework preserves the interpretability and
auditability of each component. This design principle, using structured
intermediate representations to bridge perception and reasoning, may
generalize to other safety-critical applications where VLMs must produce
quantitatively grounded judgments from visual
input~\cite{Baum2024VLMtransport, TianDriveVLM2024}.

\subsection{Depth Inference Limitations}
\label{sec:depth_limitations}

Monocular depth estimation from single RGB images is inherently
ambiguous, as identical two-dimensional projections can arise from
arbitrarily scaled three-dimensional scenes~\cite{Eigen2014,
Saxena2005, Lindell2020}. The VLM-based categorical depth scheme
adopted here is a practical compromise that acknowledges this limit
while still providing operationally useful severity stratification.

The three-class scheme (low / medium / high) was chosen to align with
established pavement evaluation
standards~\cite{FHWA2013DistressManual, OntarioReg2002} rather than to
approximate continuous metric depth. Severity judgment from a single image is
difficult for people as well: the first panel agreed at mean pairwise
$\kappa = 0.453$, and the second, prediction-free panel of
Section~\ref{sec:v1} at $\kappa = 0.605$ over 102 potholes.

Against the prediction-free panel the pipeline's severity agreement is
$\kappa = 0.139$ (weighted $\kappa = 0.320$), far below the annotators'
agreement with one another (0.605). Against physically measured depth on an
RGB-D dataset it is $\kappa = 0.098$. These two independent references agree.
The first panel ($\kappa = 0.321$ for the released labels, 0.479 for the
validation run) is the outlier, and it is the panel whose annotation sheet
displayed the models' predictions. The engine's severity classes should
therefore not be described as comparable to human judgment.

That interface choice mattered. On the same 42 potholes, removing the displayed
predictions raised agreement among annotators from 0.453 to 0.759 and lowered
the engine's agreement with the consensus from 0.321 to 0.218. The two
quantities moved in opposite directions, which is the expected signature of
anchoring. We therefore recommend prediction-free annotation interfaces for
this kind of validation. Stereo~\cite{Strehlow2024}, LiDAR, and temporal
multi-frame cues~\cite{Ming2021} are the natural avenues for moving beyond the
categorical regime.

\subsection{Limitations}
\label{sec:limitations}

Before listing specific limitations we restate the scope of the claims. The
system is an \emph{offline} infrastructure-assessment and warning-generation
framework evaluated on recorded imagery. The generated maneuver
recommendations---lane adjustments, steering offsets, and speed
reductions---are research outputs intended as an initial step toward
context-aware advisory generation. They are produced without observation of
surrounding traffic, adjacent-lane occupancy, vehicle dynamics, road friction,
or available free space, and they have not been validated in
human-factors studies, simulation, or closed-loop vehicle testing. They must
not be interpreted as executable driving guidance. Establishing that such
advisories are safe and useful to a driver is precisely the validation we
identify as the necessary next step.

Several further limitations should be acknowledged. First, all experiments
were conducted on a single public dataset collected from South
African roads~\cite{nienaber2015detecting}. The dataset covers
varied pavement, lighting, and damage conditions, but generalization
to other geographic regions and climates (e.g., freeze--thaw pothole
morphologies in northern climates) has not been empirically validated. We
tested severity classification and segmentation on one additional RGB-D
dataset (Section~\ref{sec:new_analyses}), but at close range rather than from a
dashcam viewpoint.

Second, this work evaluates the \emph{assessment} of already-detected
potholes and is not an end-to-end detection system; detection performance is
neither claimed nor measured here. The bounding boxes are dataset
ground-truth annotations used as a stand-in for an upstream detector, which
in our own prior work supplies these prompts~\cite{AbuRaddaha2024}, so that
the evaluation can isolate the contribution of geometric grounding. In a deployed
real-time system an upstream detector such as YOLO or
Faster~R-CNN~\cite{Redmon2016YOLO, Ren2015FasterRCNN} would supply
these prompts, and established detectors already achieve high
precision on pothole localization~\cite{ORDDC2024}. Real-time
deployment is not the focus of the present paper; integrating an
upstream detector is therefore an engineering step for future
deployment rather than a constraint on the methodology.

Third, the severity classes are perceived visual categories, not
measurements. Against physically measured depth they preserve the order of
depths but agree only weakly with the measured classes and under-call the
deepest potholes (Section~\ref{sec:metric_validation}); against a
prediction-free human panel they also agree only weakly
(Section~\ref{sec:v1}). Real depth from stereo, LiDAR, or depth cameras is the
most important extension.

Fourth, the geometric features are measured in the image domain and carry no
physical scale. The distance bands are image-row categories without camera
calibration (the warning prompt of Appendix~\ref{appendix:hybrid_prompt} gives
them nominal meter ranges that have not been validated), and the wheel-path
zones assume a straight road, a fixed camera, and a vehicle centered in its
lane, without modeling camera pose, lane boundaries, road curvature, vehicle
width, or lateral vehicle position.

Fifth, the risk-score weights and boundaries are design choices. They are
stable in the ranking they induce (Section~\ref{sec:risk_sensitivity}), but
no outcome data exist with which to test whether the score tracks vehicle
damage or the urgency of a maneuver.

Sixth, the human studies are small (50 scenes for the warnings and 102
potholes for severity), and the raters were volunteers rather than trained
pavement inspectors. The urgency comparison and the prompt ablation are scored
against the pipeline's own references rather than against independently
created hazard and maneuver labels, so they show which inputs change the
output, not which output is correct.

Seventh, the VLMs are hosted models. Low temperature reduces but does not
remove run-to-run variation (Section~\ref{sec:variability}), and provider-side
updates may change outputs, so exact reproduction requires the released
outputs rather than a re-run.

\subsection{Future Work}
\label{sec:future_work}

Several directions follow naturally from the findings and limitations
above.

\textbf{Temporal Fusion.} The framework currently operates on single
frames. Temporal fusion from consecutive dashcam frames could
maintain consistent risk tracking as potholes enter, traverse, and
exit the field of view, and multi-frame depth estimation via
structure-from-motion or learned temporal aggregation could upgrade
categorical depth to approximate metric values without additional
sensors.

\textbf{Vision-Language-Action Models.} The ENPAD dataset, with its
structured per-pothole risk scores and natural-language maneuver
recommendations, provides a natural training corpus for VLA
models~\cite{Brohan2023RT2} that directly map visual observations
and geometric context to driving actions.

\textbf{Trajectory Planning Integration.} The structured risk output
of the current framework could feed path-planning algorithms that
generate evasive trajectories, speed adjustments, or lane-change
decisions informed by quantitative severity rather than binary
obstacle detection~\cite{ShalevShwartz2017RSS}, closing the loop
between hazard assessment and vehicle response.

\textbf{Real-Time Deployment.} Replacing SAM~ViT-H with efficient
alternatives such as TinySAM~\cite{Chen2024TinySAM}, MobileSAM, or
FastSAM, combined with a distilled depth classifier trained on the
categorical outputs of the current system, could bring per-frame
processing below the 100\,ms threshold required for real-time
operation at typical driving speeds.

\textbf{Independent Validation of Warnings.} Independently created hazard
and maneuver labels, together with simulator or on-road human-factors studies,
are needed to establish whether the warnings are correct and safe, and not only
preferred by raters.

\textbf{Multi-Hazard Extension.} Cross-domain transfer to other
infrastructure defects (cracking, debris, surface flooding, missing
lane markings) would test the generality of the geometric grounding
approach. The modular architecture facilitates such extensions,
requiring only modified geometric feature extraction and adapted
risk-scoring parameters while retaining the same VLM reasoning and
warning generation stages.

\section{Computational Cost Analysis}
\label{sec:cost}

Experiments were conducted on a workstation equipped with dual
NVIDIA~RTX~6000~Ada GPUs (49{,}140\,MiB VRAM each) and 64\,GB system
RAM, with VLM calls routed through the Google~AI API.
Table~\ref{tab:cost_breakdown} reports the per-stage breakdown.

\begin{table}[!t]
\caption{Per-Scene Processing Time and Cost Breakdown}
\label{tab:cost_breakdown}
\centering
\setlength{\tabcolsep}{4pt}
\renewcommand{\arraystretch}{1.2}
\begin{tabular}{l c c c}
\toprule
\textbf{Stage} & \textbf{Compute} & \textbf{Time (s)} & \textbf{Cost (\$)} \\
\midrule
1: Perspective Transform  & CPU  & $<$0.01     & ---     \\
2: SAM Segmentation       & GPU  & $\sim$3.20  & ---     \\
3: Geometric Extraction   & CPU  & $\sim$0.05  & ---     \\
4: Depth + Risk Scoring   & API  & $\sim$13.30 & $\sim$0.080 \\
5: Warning Generation     & API  & $\sim$3.50  & $\sim$0.025 \\
\midrule
\textbf{Total per scene}  &      & $\sim$20.1  & $\sim$0.105 \\
\bottomrule
\end{tabular}
\end{table}

End-to-end processing averages 18--23\,s per scene, with over 83\%
attributable to API communication: Stage~4 dominates at $\sim$13.3\,s
per scene (averaging 4.8\,s per pothole at 2.76~potholes per scene),
and Stage~5 adds $\sim$3.5\,s in a single per-scene call. Local
stages collectively require $\sim$3.3\,s. Processing the full
1,119-scene dataset required 8.2\,h of wall-clock time (measured from the Stage-4 run log, 2025-12-10 15:21--23:35, i.e.\ 26.5\,s per scene including API queueing and retries) at a total API cost
of \$117.50 ($\sim$\$0.105 per scene).

At this rate an offline survey of 10{,}000~scenes would cost roughly
\$1{,}050 in API fees. The modular design
permits selective re-execution, and on-device distillation of the
API-bound stages (Section~\ref{sec:future_work}) is a natural route to
further cost and latency reduction.

\section{Conclusion}
\label{sec:conclusion}

This paper presented a five-stage modular framework that couples
deterministic geometric measurement with vision-language reasoning for the
offline risk assessment of already-detected potholes. Dashcam imagery is
processed through perspective transformation (Stage~1), zero-shot SAM
segmentation (Stage~2), six-dimensional geometric feature extraction
(Stage~3), Gemini~2.5~Pro severity classification with multi-factor
composite risk scoring (Stage~4), and multimodal hybrid warning generation
via Gemini~2.5~Pro (Stage~5). Structured input--output contracts between
stages enable isolated validation and targeted component replacement.

Across 3{,}092~assessed potholes in 1{,}119~scenes, the results
separate what geometric grounding achieves from what it does not. It changes
the warnings: relative to the image-only baseline, the hybrid method moved
26.7\% of scenes out of the high-urgency tier and raised critical-tier
assignments by 109\% (Stuart--Maxwell $\chi^{2} = 260.1$, $p < 10^{-15}$),
and in a blind study three raters preferred its warnings 3.5:1 and rated
them higher on clarity ($+0.83$, 95\%~CI $[0.54, 1.11]$) and safety
relevance ($+0.73$, $[0.36, 1.11]$) on a five-point scale. A six-arm
ablation attributes the hybrid prompt's rule-adherence to the per-pothole measurements rather than to the
rubric or worked examples. The severity classes, by contrast, agree
only weakly with independent references: $\kappa = 0.139$ against a
prediction-free three-annotator panel whose members agree with one another
at $\kappa = 0.605$, and $\kappa = 0.098$ against physically measured
depth, although the predicted classes preserve the order of measured depths.
The released ENPAD annotations are therefore a resource for developing and
comparing methods rather than a benchmark of physical severity, and the
generated maneuvers are advisory research outputs that have not been
validated for vehicle control.

Real depth sensing, independently created hazard and maneuver labels, and
human-factors validation are the most important next steps, alongside
lightweight segmentation and temporal fusion for on-vehicle use, as detailed
in Section~\ref{sec:future_work}.

\section*{Ethics Statement}
The human studies asked adult volunteers to rate images of road surfaces and
machine-generated warning text. The volunteers were acquaintances of the
authors who took part without payment; they were told how their ratings would
be used and agreed to their anonymous use in publications. No personal,
biometric, or health data were collected, the tasks posed no risk beyond
ordinary computer use, and the released data contain no information that
identifies a rater. Formal review by an institutional review board was not
sought for these rating tasks.

\section*{Data and Code Availability}
The code, the complete ENPAD annotations (Stage-4 outputs for all 3,092
assessed potholes, and risk scores and both warning sets for all 1,119 scenes),
the anonymized annotations and ratings of the three human studies, the scripts
and result files of every analysis reported in this paper, and a 30-scene image
sample are available at
\url{https://github.com/malekaburaddaha/Pothole_Assessment_Pipeline}. Code is
released under the MIT license, and annotations and human-study data under
CC~BY~4.0. The underlying imagery remains subject to the license of the
original dataset~\cite{nienaber2015detecting}; the perspective-transformed
images can be regenerated with the released Stage-1 code.
Model identifiers and access dates are:
\texttt{gemini-2.5-pro} and \texttt{gemini-2.5-flash} (Google AI API, accessed
10--15 December 2025) and \texttt{gpt-5} (OpenAI API, accessed December 2025).
Generation temperature was 0.1 for depth inference; for warning generation,
the image-only baseline used 0.3 and the hybrid prompt the API's default
sampling settings (Section~\ref{sec:ablation}). Across the 3,092-pothole depth run, 0 responses were blocked by
safety filters and all parsed successfully; in the 1,119-scene warning run all
responses parsed under the enforced JSON schema. The default outputs described
in Section~\ref{sec:methodology} were therefore never applied. The prompt
ablation of Section~\ref{sec:prompt_ablation} used 300 scenes drawn uniformly at
random (seed~5); the depth panels and the warning study used the samples
described in Sections~\ref{sec:depth_validation}, \ref{sec:human_eval},
and~\ref{sec:v1}. Because these are hosted
models, exact reproducibility cannot be guaranteed: a low temperature reduces
but does not eliminate run-to-run variation, and provider-side model updates
may alter outputs. We therefore release the complete generated outputs
alongside the code.

\section*{Acknowledgment}
The authors thank the human annotators who participated in the depth
classification and warning evaluation studies. An AI-based tool was used to
refine the visual layout of the framework block diagram (Fig.~\ref{fig:pipeline})
so that it is easier to read; the authors designed the diagram and verified its
content.

\appendix

\section{VLM Prompt Templates}
\label{appendix:prompts}

This appendix provides the prompt templates used in this work in condensed
form; the complete prompts, exactly as they were sent to the models, are in the
released code. Where a prompt states physical
quantities, such as the depth bands or the meter ranges attached to the
distance categories, these are nominal descriptions given to the model, not
calibrated measurements (Sections~\ref{sec:stage3}
and~\ref{sec:metric_validation}).
Appendix~\ref{appendix:depth_prompt} gives the depth inference prompt
used in Stage~4. Appendix~\ref{appendix:warning_prompts} provides both
warning generation prompts used in the ablation study of
Section~\ref{sec:ablation}: an image-only baseline and the multimodal
hybrid variant.

\subsection{Stage 4 --- Depth Inference Prompt}
\label{appendix:depth_prompt}

The following template is submitted once per pothole to Gemini 2.5 Pro.
Placeholders in curly braces are populated at runtime from the geometric
feature vector of Stage~3 (Table~\ref{tab:features}). The conditional
dual-image note in the final block is appended only when the pothole is
classified as \emph{far} or its crop dimension falls below 100\,pixels
(see Section~\ref{sec:stage4}).

\begin{quote}
\footnotesize\ttfamily\raggedright\sloppy
\noindent\textbf{[SYSTEM CONTEXT]}~Scientific analysis of road
infrastructure for autonomous vehicle safety. Images contain inanimate
road surfaces only.

\medskip
\noindent You are an expert computer vision system assessing road hazards
for autonomous vehicles.

\medskip
\noindent\textbf{TASK:} Analyze the provided pothole image to determine
its DEPTH category.

\medskip
\noindent\textbf{Known Geometric Properties (measured via computer vision):}
\begin{itemize}[noitemsep, topsep=2pt]
    \item Area: \{area\} pixels\textsuperscript{2}
    \item Circularity: \{circularity\} \hspace{1em}(0 = irregular,\; 1 = perfect circle)
    \item Solidity: \{solidity\} \hspace{1em}(ratio of area to convex hull)
    \item Aspect Ratio: \{aspect\_ratio\}
    \item Distance Category: \{Near\,/\,Medium\,/\,Far\}
    \item Lateral Position: \{in\_path\_left\,/\,right\;|\;\allowbreak in\_lane\_far\_left\,/\,right\}
\end{itemize}

\medskip
\noindent\textbf{Depth Category Definitions:}
\begin{itemize}[noitemsep, topsep=2pt]
    \item \textbf{low:} Shallow surface depression ($<$2\,cm depth)
    \item \textbf{medium:} Moderate depth pothole (2--5\,cm depth)
    \item \textbf{high:} Deep, hazardous pothole ($>$5\,cm depth)
\end{itemize}

\medskip
\noindent\textbf{Your Analysis:} Based on visual evidence in the image
(shadow depth, edge sharpness, surface texture, darkness/lighting),
classify this pothole's DEPTH into ONE category: low, medium, or high.

\medskip
\noindent\textbf{Visual Cues to Consider:}
\begin{enumerate}[noitemsep, topsep=2pt]
    \item Shadow depth and darkness \hspace{1em}(deeper $\rightarrow$ darker shadows)
    \item Edge definition and sharpness \hspace{1em}(deeper $\rightarrow$ sharper edges)
    \item Surface texture visibility \hspace{1em}(deeper $\rightarrow$ less visible bottom texture)
    \item Lighting contrast between pothole interior and road surface
\end{enumerate}

\medskip
\noindent\textit{[Appended when dual-image input is used:]}
Two images are provided: (1)~Full road scene for spatial context and
lighting reference; (2)~Cropped pothole region for depth assessment.
Use the full scene to understand overall lighting and road texture,
then focus on the cropped region for depth characteristics.

\medskip
\noindent\textbf{Response Format:} Respond ONLY with valid JSON:
\{"depth\_category": "low|medium|high",
"confidence": "low|medium|high",
"reasoning": "brief explanation of visual cues used"\}
\end{quote}

\subsection{Stage~5 --- Warning Generation Prompts}
\label{appendix:warning_prompts}

The two prompts below correspond to the ablation variants evaluated in
Section~\ref{sec:ablation}. Both target the same six-field JSON output
schema but differ in the information modalities provided to the model.

\subsubsection{Image-Only Baseline Prompt}
\label{appendix:image_only_prompt}

This prompt provides only the RGB scene image; no geometric features,
risk scores, or depth categories are supplied. The system context preamble
serves to prevent content-filter rejections of road damage imagery.

\begin{quote}
\footnotesize\ttfamily\raggedright\sloppy
\noindent You are an AI assistant supporting academic research on
autonomous vehicle safety systems. Examine this road scene image for
potholes and pavement defects. Based ONLY on what you observe in the
image, generate a safety warning.

\medskip
\noindent\textbf{Assessment Guidelines:}
\begin{itemize}[noitemsep, topsep=2pt]
    \item \textbf{urgency\_level:} critical (large, deep damage requiring
    immediate avoidance) $|$ high (significant defect, avoidance strongly
    recommended) $|$ medium (moderate deterioration, caution advised) $|$
    low (minor defect, awareness only)
    \item \textbf{warning\_message:} Clear, professional, 1--2 sentences
    \item \textbf{maneuver\_suggestion:} Specific driving action
    (speed reduction, lane adjustment)
    \item \textbf{reasoning:} Explain visual observations
    (size, depth appearance, position)
\end{itemize}

\medskip
\noindent Output as JSON with 6 fields: \texttt{warning\_message},
\texttt{maneuver\_suggestion}, \texttt{urgency\_level},
\texttt{reasoning}, \texttt{affected\_lanes},
\texttt{recommended\_speed\_reduction}.
\end{quote}

\subsubsection{Multimodal Hybrid Prompt}
\label{appendix:hybrid_prompt}

This prompt supplies the RGB scene together with the full structured
assessment: per-pothole depth categories, geometric features, risk scores,
and a worked-example decision rubric. Scene-level placeholders
(scene ID, pothole count, per-pothole attribute rows) are populated
at runtime from the Stage~4 risk scoring output. The distance ranges and
reaction times in the domain knowledge reference are nominal values supplied to
the model; the distance categories themselves are uncalibrated image-row bands
(Section~\ref{sec:stage3}).

\begin{quote}
\footnotesize\ttfamily\raggedright\sloppy
\noindent You are an expert autonomous vehicle safety advisor analyzing
a road scene image along with quantified pothole assessments.

\medskip
\noindent\textbf{DOMAIN KNOWLEDGE REFERENCE:}

\smallskip
\noindent\textit{Depth Categories:}
Low ($<$2\,cm): minor surface defect, safe at reduced speed $|$
Medium (2--5\,cm): moderate hazard, can damage tires/suspension $|$
High ($>$5\,cm): severe pothole, high risk of tire blowout or loss of
control.

\smallskip
\noindent\textit{Distance Categories (reaction time):}
Near (0--10\,m): $<$1\,s at 30\,mph, immediate action required $|$
Medium (10--20\,m): 1--2\,s, controlled braking/steering possible $|$
Far ($>$20\,m): $>$2\,s, ample time for smooth maneuvers.

\smallskip
\noindent\textit{Lateral Position:}
in\_path\_left/right: direct wheel path, highest collision probability $|$
in\_lane: avoidable by slight steering $|$
lane\_boundary: peripheral, low risk.

\smallskip
\noindent\textit{Risk Score:}
0--40 (Low) $|$ 40--60 (Medium) $|$ 60--80 (High) $|$ 80--100 (Critical).

\medskip
\noindent\textbf{DECISION-MAKING RUBRIC:}

\noindent CRITICAL: High depth + Near distance + In wheel path,
OR Risk $\geq$ 80.

\noindent HIGH: Medium/high depth + Near distance, OR Risk 60--79.

\noindent MEDIUM: Low/medium depth + Near, OR High depth + Far,
OR Risk 40--59.

\noindent LOW: Low depth + Medium/far distance, OR Risk $<$40.

\medskip
\noindent\textbf{WORKED EXAMPLES:} [Four complete scenarios provided
at runtime, demonstrating critical, high, medium, and low urgency
classifications with reasoning citing specific numerical values.]

\medskip
\noindent\textbf{CURRENT SCENE ASSESSMENT:}

\noindent Scene ID: \{scene\_id\} $|$ Total Potholes: \{n\} $|$
Critical/High Risk Count: \{k\}

\noindent Pothole 1: Depth: \{category\} (\{confidence\} confidence) $|$
Size: \{area\}\,px\textsuperscript{2} $|$ Distance: \{dist\} $|$
Lateral: \{position\} $|$ Risk: \{score\}/100 (\{severity\})

\noindent [Additional pothole rows follow the same format\ldots]

\medskip
\noindent Use BOTH visual context AND quantified metrics to generate the
warning. Cite specific risk scores in the reasoning field. Output as
JSON with the same 6 fields as above.
\end{quote}

\bibliographystyle{IEEEtran}
\bibliography{references}

\begin{IEEEbiography}[{\includegraphics[width=1in,height=1.25in,
clip,keepaspectratio]{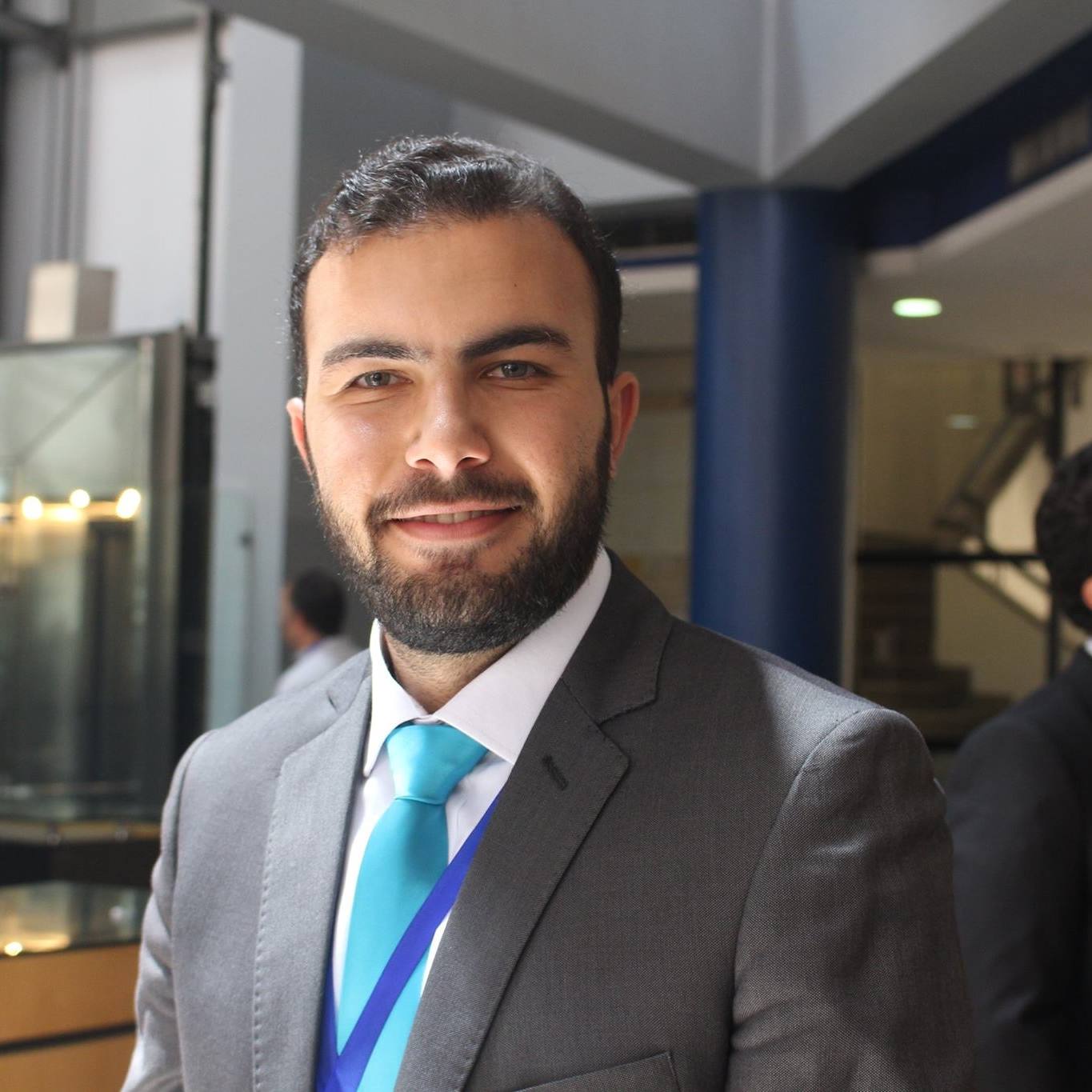}}]{Abdalmalek Aburaddaha}
received the B.Sc.\ degree in automotive engineering from Al-Balqa' Applied
University, Amman, Jordan, in 2017, the M.Sc.\ degree in computer engineering
from the University of Michigan--Dearborn, Dearborn, MI, USA, in 2023, and is
currently pursuing the Ph.D.\ degree in computer engineering at the same
institution. He served as a Teaching Assistant with the University of
Michigan--Dearborn from 2021 to 2026 and is currently a Research Assistant with
the same university.
He has authored and coauthored several peer-reviewed publications in
journals and conferences spanning computer vision, autonomous driving,
and multimodal sensing. His research interests include deep learning for
autonomous vehicle perception, vision-language and action models and foundation models
for safety-critical AI, hard-example mining, and the application of agentic AI and large language models to structured perception and actionable decision-making.
\end{IEEEbiography}

\begin{IEEEbiography}[{\includegraphics[width=1in,height=1.25in,
clip,keepaspectratio]{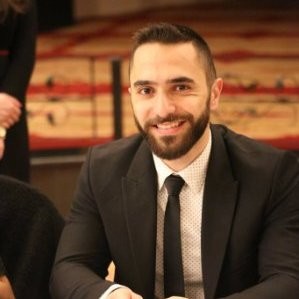}}]{Zaid A. El Shair}
received the B.Sc. degree in Computer Engineering from Princess Sumaya University for Technology, Amman, Jordan, in 2018, and the M.S.E. degree in Computer Engineering and the Ph.D. degree in Electrical, Electronics, and Computer Engineering from the University of Michigan--Dearborn, Dearborn, MI, USA, in 2019 and 2024, respectively. He was a Graduate Student Instructor with the University of Michigan--Dearborn from 2018 to 2023 and served as a Research Associate and Postdoctoral Research Fellow from 2023 to February 2025. He also completed internships with Ford Motor Company in 2020 and 2021. He is currently a Machine Learning Engineer with Torc Robotics, where he develops machine learning and autonomous perception algorithms and researches methods for accelerating and deploying them efficiently on embedded and high-performance computing hardware platforms. He has authored and coauthored several peer-reviewed publications and continues to actively collaborate on and publish research in computer vision, machine learning, autonomous perception systems, and embedded AI. His research interests include computer vision, multimodal perception, autonomous systems, deep learning-based perception methods, hardware-accelerated AI systems, and event-based vision for object detection, tracking, and related applications.
\end{IEEEbiography}

\begin{IEEEbiography}[{\includegraphics[width=1in,height=1.25in,
clip,keepaspectratio]{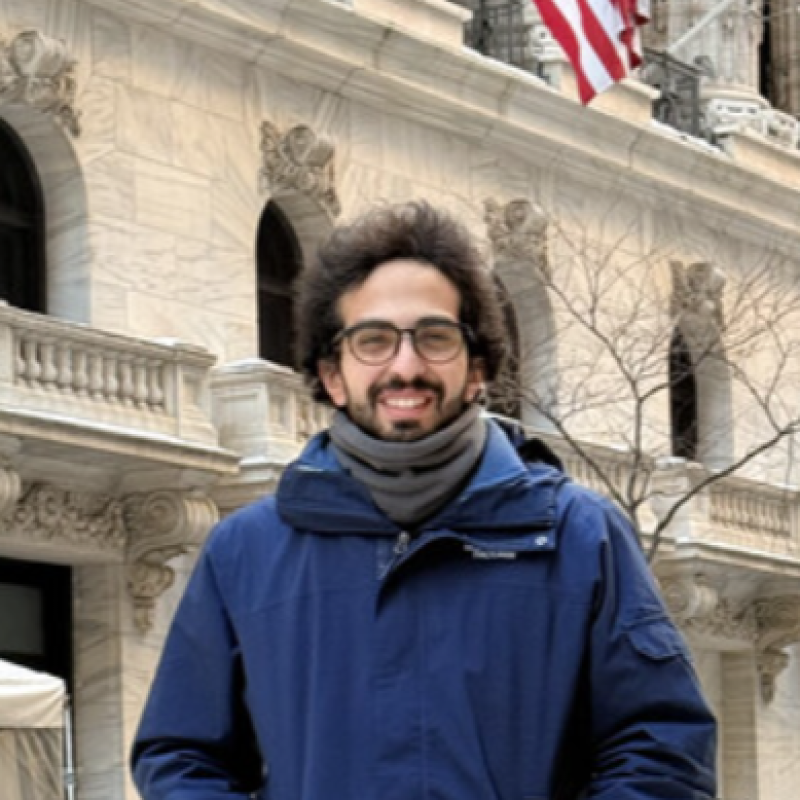}}]{Ali Al-Bustami}
received the B.Sc.\ degree in mechatronics engineering from The University of
Jordan, Amman, Jordan, and is currently pursuing the M.S.\ degree in computer
engineering at the University of Michigan--Dearborn, Dearborn, MI, USA, where
he has also served as a Research Assistant. He has authored and coauthored
several peer-reviewed papers. His research interests include machine learning,
computer vision, robotics, and AI systems for autonomous and real-world
applications.
\end{IEEEbiography}

\begin{IEEEbiography}[{\includegraphics[width=1in,height=1.25in,
clip,keepaspectratio]{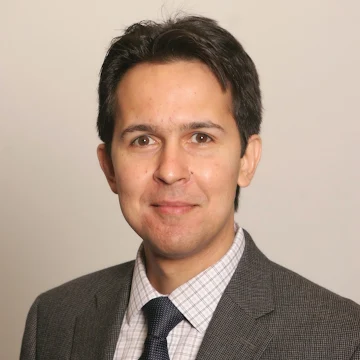}}]{Samir A. Rawashdeh}
(Senior Member, IEEE) received the B.Sc.\ degree in electrical engineering from The University of Jordan, Amman, Jordan, in 2007, and the M.S.E.E.\ and Ph.D.\ degrees in electrical engineering from the University of Kentucky, Lexington, KY, USA, in 2009 and 2013, respectively. In 2014, he joined the Department of Electrical and Computer Engineering, University of Michigan--Dearborn, Dearborn, MI, USA, where he is currently a Full Professor. His research interests include robot perception, embedded systems, and smart health.
\end{IEEEbiography}

\vfill

\end{document}